%% file: main.tex
\documentclass{article}
\usepackage{colm2024_conference}

\usepackage{microtype}
\usepackage{hyperref}
\usepackage{url}
\usepackage{booktabs}
\usepackage{soul}
\usepackage[T1]{fontenc}
\usepackage{listings}

\definecolor{darkblue}{rgb}{0, 0, 0.5}
\definecolor{typhoonpurple}{RGB}{113, 107, 216}
\hypersetup{colorlinks=true, citecolor=darkblue, linkcolor=darkblue, urlcolor=darkblue}

\newcommand\sbullet[1][.75]{\mathbin{\vcenter{\hbox{\scalebox{#1}{$\bullet$}}}}}
\usepackage{graphicx}
\usepackage{amsmath,amssymb,amsfonts}
\usepackage{algorithmic}
\usepackage{textcomp}
\usepackage{xcolor}

\usepackage{color, tabularx}
\usepackage{booktabs, multirow}
\usepackage{float}
\usepackage{url}
\usepackage{color, colortbl}
\definecolor{Gray}{gray}{0.9}
\usepackage{float}
\usepackage{caption}
\usepackage{diagbox}
\usepackage{hyperref}

\usepackage{pifont}% http://ctan.org/pkg/pifont

\usepackage{cleveref}

\newcommand{\cmark}{\ding{51}}%
\newcommand{\xmark}{\ding{55}}%
\NewDocumentCommand\emojilogo{}{
\includegraphics[scale=0.055, trim=1cm 11.2cm 0 0]{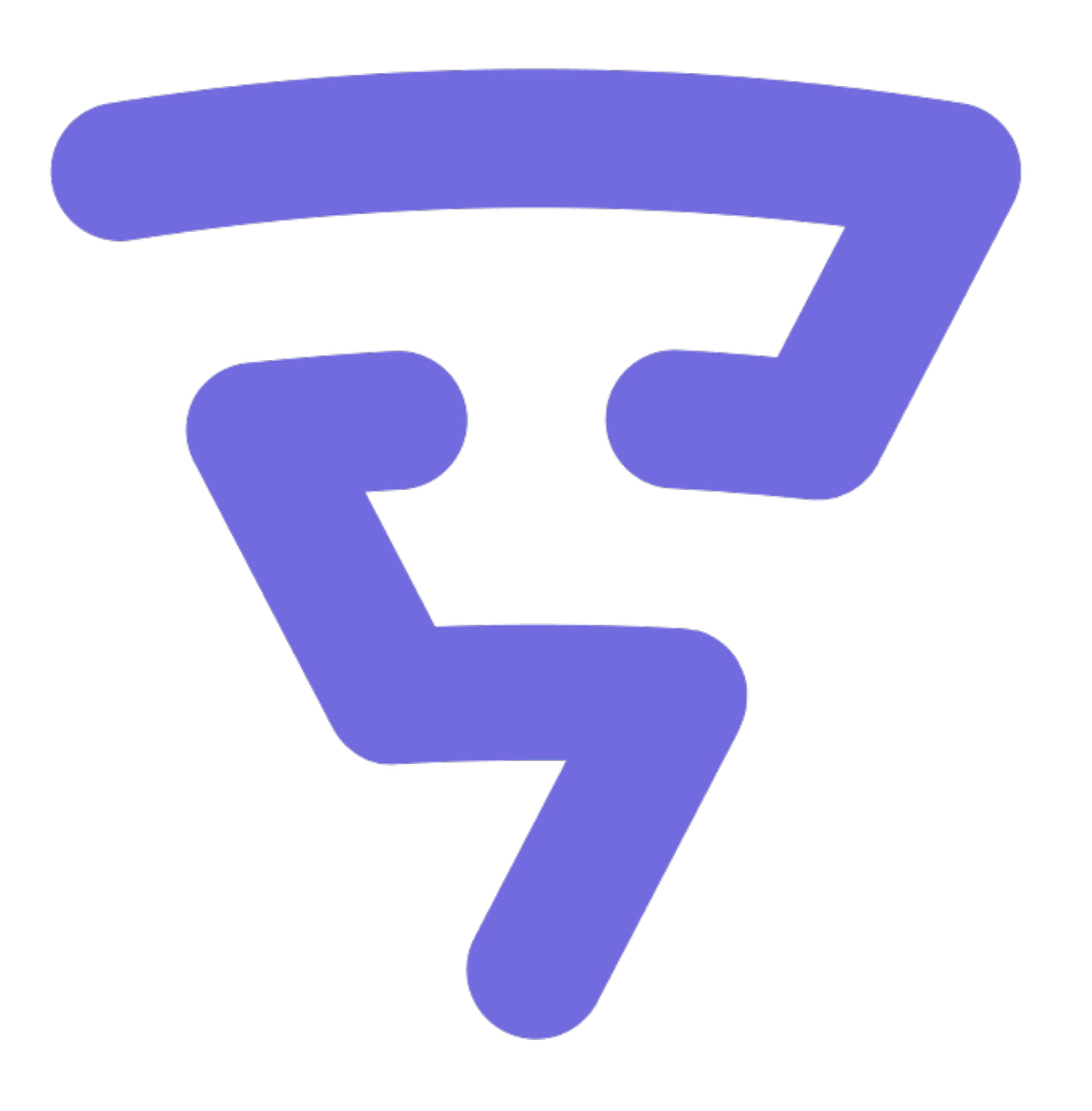}
}

\title{\emojilogo Typhoon 2: A Family of Open Text and Multimodal\\
\phantom{ABE.}Thai Large Language Models}

\author{\vspace{-4mm} \\
Kunat Pipatanakul, Potsawee Manakul, Natapong Nitarach, \vspace{0.2mm} \\
Warit Sirichotedumrong, Surapon Nonesung, Teetouch Jaknamon, \vspace{0.2mm} \\
Parinthapat Pengpun, Pittawat Taveekitworachai, Adisai Na-Thalang, \vspace{0.2mm} \\ 
Sittipong Sripaisarnmongkol, Krisanapong Jirayoot, Kasima Tharnpipitchai \vspace{5.5mm} \\
\textbf{SCB 10X, SCBX} \\
\texttt{contact@opentyphoon.ai} \\
}

\colmfinalcopy
\begin{document}

\maketitle

\begin{abstract}
% intro
This paper introduces Typhoon 2, a series of text and multimodal large language models optimized for the Thai language. The series includes models for text, vision, and audio. \textbf{Typhoon2-Text} builds on state-of-the-art open models, such as Llama 3 and Qwen2, and we perform continual pre-training on a mixture of English and Thai data. We employ post-training techniques to enhance Thai language performance while preserving the base models' original capabilities. We release text models across a range of sizes, from 1 to 70 billion parameters, available in both base and instruction-tuned variants. To guardrail text generation, we release Typhoon2-Safety, a classifier enhanced for Thai cultures and language. \textbf{Typhoon2-Vision} improves Thai document understanding while retaining general visual capabilities, such as image captioning. \textbf{Typhoon2-Audio} introduces an end-to-end speech-to-speech model architecture capable of processing audio, speech, and text inputs and generating both text and speech outputs. 

\end{abstract}

\subsection*{Summary of the Typhoon2 Models}

\renewcommand{\arraystretch}{1.25} % Adjust the row spacing (default is 1.0)

\begin{table}[!ht]
    \centering
    \small
    \tabcolsep=0.8mm
    \begin{tabular}{lll}
\toprule
 \textbf{Model}   & \textbf{Base Model}  & \textbf{Link to HuggingFace} \\
 \midrule
\rowcolor{Gray}
\multicolumn{3}{l}{\textbf{Text}}  \\
\texttt{Typhoon2-1B-Base}      & \multirow{2}{*}{\texttt{Llama-3.2-1B}}  & \href{https://huggingface.co/scb10x/llama3.2-typhoon2-1b}{\texttt{scb10x/llama3.2-typhoon2-1b}} \\ 
\texttt{Typhoon2-1B-Instruct}  &  & \href{https://huggingface.co/scb10x/llama3.2-typhoon2-1b-instruct}{\texttt{scb10x/llama3.2-typhoon2-1b-instruct}} \\ 
\texttt{Typhoon2-3B-Base}      & \multirow{2}{*}{\texttt{llama-3.2-3B}}  & \href{https://huggingface.co/scb10x/llama3.2-typhoon2-3b}{\texttt{scb10x/llama3.2-typhoon2-3b}} \\ 
\texttt{Typhoon2-3B-Instruct}  &  & \href{https://huggingface.co/scb10x/llama3.2-typhoon2-3b-instruct}{\texttt{scb10x/llama3.2-typhoon2-3b-instruct}} \\ 
\texttt{Typhoon2-7B-Base}      & \multirow{2}{*}{\texttt{Qwen2.5-7B}}  & \href{https://huggingface.co/scb10x/typhoon2-qwen2.5-7b}{\texttt{scb10x/typhoon2-qwen2.5-7b}} \\ 
\texttt{Typhoon2-7B-Instruct}  &  & \href{https://huggingface.co/scb10x/typhoon2-qwen2.5-7b-instruct}{\texttt{scb10x/typhoon2-qwen2.5-7b-instruct}} \\ 
\texttt{Typhoon2-8B-Base}      & \multirow{2}{*}{\texttt{Llama-3.1-8B}}  & \href{https://huggingface.co/scb10x/llama3.1-typhoon2-8b}{\texttt{scb10x/llama3.1-typhoon2-8b}} \\ 
\texttt{Typhoon2-8B-Instruct}  &  & \href{https://huggingface.co/scb10x/llama3.1-typhoon2-8b-instruct}{\texttt{scb10x/llama3.1-typhoon2-8b-instruct}} \\ 
\texttt{Typhoon2-70B-Base}     & \multirow{2}{*}{\texttt{Llama-3.1-70B}}  & \href{https://huggingface.co/scb10x/llama3.1-typhoon2-70b}{\texttt{scb10x/llama3.1-typhoon2-70b}} \\ 
\texttt{Typhoon2-70B-Instruct} &  & \href{https://huggingface.co/scb10x/llama3.1-typhoon2-70b-instruct}{\texttt{scb10x/llama3.1-typhoon2-70b-instruct}} \\ 
\rowcolor{Gray}
\multicolumn{3}{l}{\textbf{Safety Classifier}}  \\
\texttt{Typhoon2-Safety}  & \texttt{mdeberta-v3-base}   & \href{https://huggingface.co/scb10x/typhoon2-safety-preview}{\texttt{scb10x/typhoon2-safety-preview}} \\

\rowcolor{Gray}
\multicolumn{3}{l}{\textbf{Multimodal}}  \\
\texttt{Typhoon2-Vision} & \texttt{Qwen2-VL-7B-Instruct}  & \href{https://huggingface.co/scb10x/typhoon2-qwen2vl-7b-vision-instruct}{\texttt{scb10x/typhoon2-qwen2vl-7b-vision-instruct}} \\ 
\texttt{Typhoon2-Audio}  & \texttt{Typhoon2-8B-Instruct}  & \href{https://huggingface.co/scb10x/llama3.1-typhoon2-audio-8b-instruct}{\texttt{scb10x/llama3.1-typhoon2-audio-8b-instruct}} \\
 \bottomrule
    \end{tabular}
    \caption{Models released in the Typhoon2 series}
    \label{tab:summary}
\end{table}

\renewcommand{\arraystretch}{1.0} % Reset to default if needed for other tables

\newpage
\tableofcontents

\input{1_introduction}
\clearpage
\input{2_pretraining}
\clearpage
\input{3_posttraining}
\clearpage
\input{4_vision}
\clearpage
\input{5.1_audio_input}
\input{5.2_audio_output}
% \clearpage

\section{Conclusions}
This technical report introduced Typhoon 2, a series of Thai LLMs, comprising text models in multiple sizes (both base and instruction-tuned variants) and multimodal models for vision and audio tasks. Our evaluation demonstrates superior performance across a majority of evaluated tasks, including math and reasoning. Typhoon2-Text models have an extended context length of up to 100,000 tokens, compared to 8,192 tokens of Typhoon~1.5. Typhoon2-Text also features function calling capabilities, achieving state-of-the-art results. Additionally, we include a safety classifier that delivers state-of-the-art performance specifically for Thai.

Typhoon 2 series also includes multimodal models, focusing on vision and audio. For visual understanding, Typhoon2-Vision achieves significantly improved Thai document understanding such as Thai OCR performance compared to its predecessor, while Typhoon2-Audio has evolved into an end-to-end speech-to-speech model, capable of generating simultaneous text and speech outputs. We have made all research artifacts, including models' weights, publicly available. We hope this work will accelerate AI advancements for the Thai language and inspire further innovation in the field.

\section{Acknowledgments}
We thank Thanapong Boontaeng for his insightful advice and feedback on this technical report and the overall project. We also thank Mukaya Panich and other members of SCB~10X for their invaluable support and guidance throughout the project. We also extend our appreciation to Kaweewut Temphuwapat and the SCBX Innovation Team for their support, resources, and valuable insights that have made this project possible. Lastly, we are grateful to the global and local AI communities for open-sourcing resources and knowledge.

\bibliography{colm2024_conference}
\bibliographystyle{colm2024_conference}

\section*{Contributions}

\textbf{Typhoon Text:} Kunat Pipatanakul, Surapon Nonesung, Teetouch Jaknamon

\textbf{Typhoon Vision:} Natapong Nitarach, Surapon Nonesung, Parinthapat Pengpun, Kunat Pipatanakul

\textbf{Typhoon Audio:} Potsawee Manakul, Warit Sirichotedumrong, Kunat Pipatanakul

\textbf{Engineering \& Infrastructure \& Applications:} Sittipong Sripaisarnmongkol, Pittawat Taveekitworachai

\textbf{Data Annotation:} Adisai Na-Thalang

\textbf{Business \& Leadership:} Krisanapong Jirayoot, Kasima Tharnpipitchai

% \appendix
% \section*{Appendix}

\end{document}

%% file: 1_introduction.tex
\newpage

\section{Introduction}
Foundation models are general models which can serve as the backbone in a wide range of AI applications involving language, vision, speech, and other modalities. There are a number of widely popular language model families, including open\footnote{In this paper, we do not make a distinction between open-source and open-weights. The term \textit{open} is used for referring to both options.} families such as Llama 3 \citep{llama3}, Qwen2.5 \citep{qwen2}, Phi 3 \citep{abdin2024phi} and proprietary families, such as GPT-4o, Claude 3, Gemini 1.5. However, although these models are multilingual and can work in a range of languages, they are developed as English-centric. Hence, the community has considered enhancing the performance of open foundation models on a small set of languages for their country or region.

In South East Asia (SEA), the efforts to enhance foundation language models for SEA languages include SeaLLM \citep{damonlp2024seallm3}, SEA-LION \citep{sea_lion_2024}, and Sailor \citep{sailor}. When it comes to the Thai language, there are model families such as WangChan~\citep{charin_polpanumas_2023_7878101}, Typhoon \citep{pipatanakul2023typhoon}, OpenThaiGPT~\citep{yuenyong2024openthaigpt}, and Pathumma \citep{pathummallm}. 

To continue our commitment in advancing Thai foundation models, this work introduces a new series of state-of-the-art Thai language and multimodal models, \textbf{Typhoon2}. Following our previous releases, these models are optimized for Thai and English, building on open-source models such as Llama 3 and Qwen2.5. The text models, \textbf{Typhoon2-Text}, are improved over Typhoon 1.5 in various aspects, including data filtering techniques for pre-training, complex instruction data development for improved post-training, long context, and function calling capabilities of the models. These text models are also now available in a range of sizes, consisting of 1B, 3B, 7B, 8B, and 70B parameters. We offer both pre-trained and instruction-tuned variants for each size. In addition, we introduce \textbf{Typhoon2-Safety} a safety text classifier designed to detect Thai-sensitive content and enhance the security of LM-integrated systems.

Furthermore, the Typhoon2 series is {multimodal}. The vision model, \textbf{Typhoon2-Vision}, builds on the first Typhoon-Vision model by enhancing Thai document understanding capabilities, such as optical character recognition (OCR). The audio model, \textbf{Typhoon2-Audio}, extends the first Typhoon-Audio model, evolving into an {end-to-end} speech processing and generation model capable of understanding audio, speech, and text, while generating text and speech outputs in parallel. This report provides details and insights from our development. We make the weights of all models publicly available on Hugging Face Hub where the summary of our release is shown in Table~\ref{tab:summary}.

%% file: 2_pretraining.tex
\section{Pre-training}

This section of the report details the pre-training phase of Typhoon 2, which continues to build on the same motivation as its predecessor: to construct the highest-quality corpus that represents the Thai culture and language. Typhoon 2 builds upon the foundation laid by the previous iteration of Typhoon. The primary objective of this iteration is to develop a more diverse and high-quality dataset in Thai for pre-training. This goal is accomplished by implementing multiple data-gathering pipelines designed to target various domains and subsets within the Thai language.

\subsection{Data Source \& Typhoon1-Corpus}
This section outlines the preparation process for Typhoon 1, which can be summarized in five steps. The first four steps involve preparing the base corpus, which will serve as a reference dataset used for filtering additional data in \Cref{sec:gather}. The fifth step focuses on selecting the general Typhoon1-Corpus \citep{pipatanakul2023typhoon}.

\subsubsection{Base Corpus Preparation}

\textbf{Step1 - Scaling the Data}: We initiate the process by increasing the number of Common Crawl (\texttt{CommonCrawl}) packages compared to the previous iteration, where we process a total of 40 packs of the \texttt{CommonCrawl} data.

\textbf{Step2 - Text Extraction}: Although \texttt{CommonCrawl} provides a pre-extracted \texttt{WET} subset, several studies suggest that WET files exhibit lower quality compared to \texttt{WARC} files extracted using external tools such as Trafilatura \citep{dclm,refinedweb}. In our study, we begin with a total of 40 packs of Thai CommonCrawl data with a cut-off date of September 2023 and perform HTML extraction from scratch using Trafilatura\footnote{\url{https://trafilatura.readthedocs.io/en/latest/}}. This process results in a significantly larger dataset, comprising approximately 3 TB of Thai text, or approximately 200 billion Llama3 tokens.

\textbf{Step3 - Strict Deduplication}: To ensure data quality and minimize redundancy, we implement a fuzzy deduplication pipeline using MinHash and locality-sensitive hashing (LSH) algorithms.\footnote{\url{https://github.com/ChenghaoMou/text-dedup}}

\textbf{Step4 - Heuristic Filtering}: To filter out pages containing excessive search engine optimization (SEO) content, very short documents, and low-quality texts, we evaluate each line using signals such as the ratio of numbers to text and the punctuation density. At the document level, document filtering is done based on other metrics, including newline ratios and overall document length. After deduplication and heuristic filtering, we have approximately 44B tokens of Thai text.

\subsubsection{Typhoon 1 General Corpus}
To ensure that the pre-trained data accurately represents the Thai language and culture, we employ a filtering process on the base corpus, involving a human-in-the-loop at the domain level in the same manner as Typhoon \citep{pipatanakul2023typhoon}. This approach ensures that the content is both relevant and appropriate, preserving its authenticity. Consequently, we obtain 5 billion high-quality Thai tokens, which serve as the foundation for training the original Typhoon and Typhoon 1.5 models.

\subsection{Gathering Diverse and High-Quality Thai Documents}\label{sec:gather}
To enhance our dataset, we observed a growing trend in pre-training large language models (LLMs) that emphasizes sourcing high-quality data from general corpora. Following this approach, we develop multiple pipelines to collect documents from a range of diverse domains, focusing on high-quality content absent from our existing general corpus.

\newpage
Our approach is aimed to address two key questions:
\begin{enumerate}
    \item \textbf{What represents ``Thai"?} – We curate data encapsulating Thai cultural knowledge.
    \item \textbf{What defines a ``state-of-the-art" LLM?} – We follow the global trend of filtering the web corpus to gather high quality \& high educational text \citep{fineweb,dclm,llama3}.
\end{enumerate}

As a result, we augment our dataset with an additional 12B high-quality Thai tokens through this process.

\subsubsection{Culturally Relevant Thai Text}

A tailored methodology for content collection is developed to address cultural nuances in the text data, drawing upon principles of the fine-web educational approach \citep{fineweb}. Specifically, we use an LLM to annotate 50,000 randomly selected entries from the base corpus. Each entry is assessed for its cultural relevancy and educational value in understanding Thai culture, using a scale ranging from 1 to 5. Next, we fine-tune the classification head of BGE-M3~\citep{bgem3} using the labeled dataset obtained from the previous process. We employ a smaller model to predict the Thai cultural value of samples within our base corpus. Subsequently, we filter only those with a predicted value of 4 or higher and use these instances to train Typhoon 2.

\subsubsection{High-Quality Text Selection}
Inspired by DCLM \citep{dclm} and the importance of high-quality text filtering in Thai, we develop a \texttt{fastText} \citep{fasttext} classifier tailored for the Thai language. Given the low-resource nature of Thai, we conduct multiple iterations of training to refine our classifier. First, we compile an instruction-following dataset in Thai, drawing from WangchanThaiInstruct~\citep{wangchaninstruct} and samples from our Typhoon Instruct dataset~\citep{pipatanakul2023typhoon} as positive examples, while incorporating random examples, including toxic content and junk websites, as negative examples. Next, we apply the classifier developed in the initial iteration to classify approximately 200,000 samples from a base corpus. After manual inspection and filtering, we train another iteration of the fastText classifier. This iterative approach results in a high-quality Thai text classifier capable of filtering content with a predicted quality ratio exceeding 0.5 as a high-quality sample.

\subsubsection{Synthetic Textbook}
To address the lack of textbooks in our general corpus, we employ a method inspired by the Phi and Cosmopedia \citep{cosmopedia,phi1} approach. We use LLM to create an augmented dataset of 5,000 text samples with styles that emulate textbooks, blog posts, and academic materials. This augmented dataset is then used to fine-tune \texttt{typhoon-1.5-8b-instruct} on a text augmentation task involving raw text and style information. The fine-tuned model is used to augment 20\% of our general corpus to resemble textbook-style content.

\subsubsection{High-Educational Content}
To gather high educational content in Thai, we use a similar approach as in collected culture-related text in Thai and \texttt{fine-web edu} \citep{fineweb}. Specifically, an LLM is used to annotate 50K samples and we fine-tune the BGE-M3 classification head. We filter only with a predicted value of 3 or higher due to the low level of educational content.

\subsubsection{Other High-Quality Sources}
We also incorporate highly educational sources, such as Thai Wikipedia, into our final pre-training dataset.

\subsection{Data Mixture}
To ensure good final model performance, it is imperative to determine the proportion of various data sources within the pre-training dataset. Since our approach involves continual pre-training (CPT), maintaining an awareness of the original pre-trained distribution is crucial.

We conduct a series of experiments to optimize the final model's performance. In the first stage, we independently evaluate each new data source to ensure that individual subsets contribute positively. We examine this by performing CPT on the 1.5B Qwen2.5~\citep{qwen2} model using a similar recipe to \citet{datasparkjoy} and verify that each of the data sources improves one of the metrics or scores based on M3Exam \citep{m3exam} and/or ThaiExam \citep{pipatanakul2023typhoon}.

\begin{figure}[!ht]
    \centerline{
\includegraphics[width=0.9\linewidth,keepaspectratio]{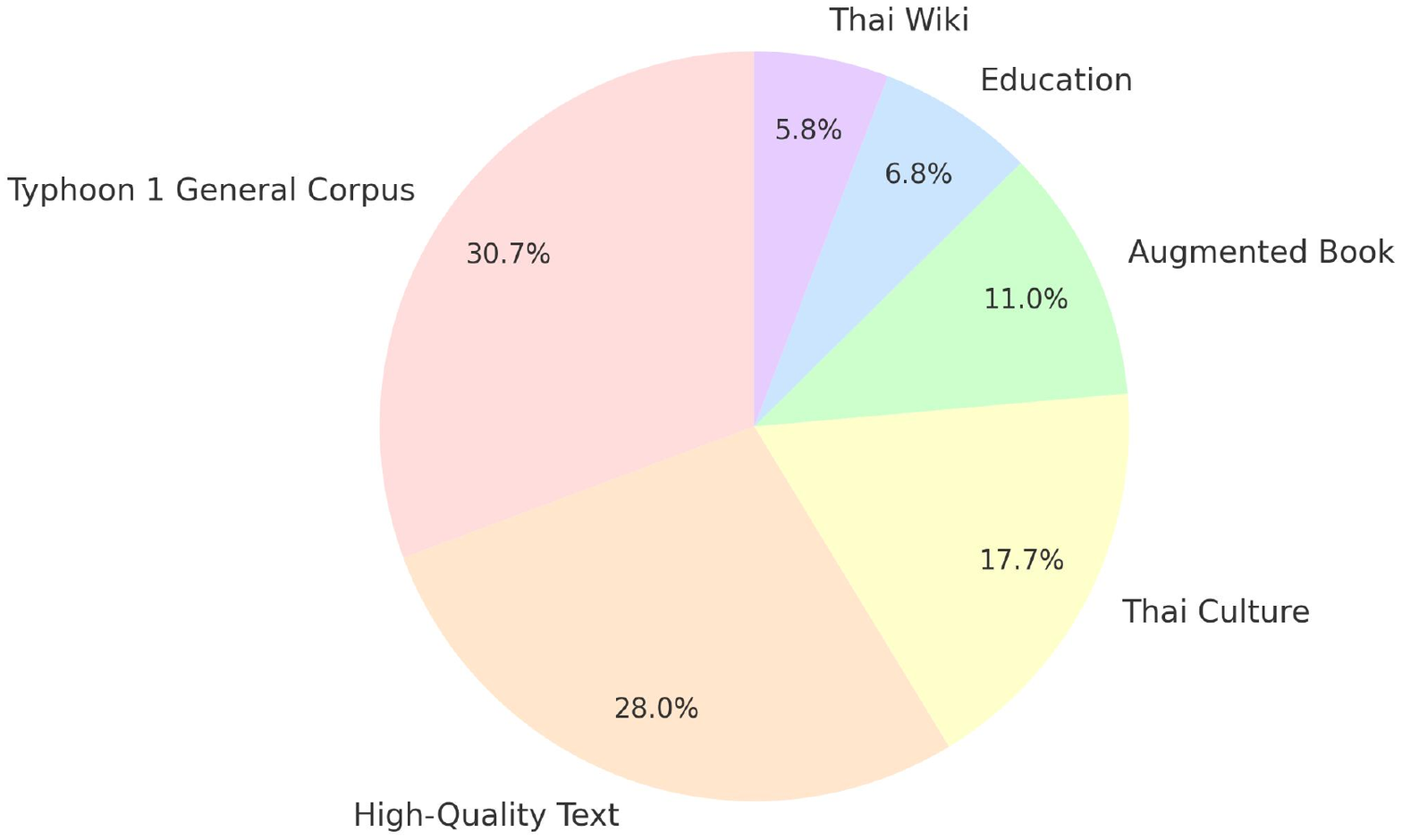}}
    \caption{Thai Pretraining Data Mixture}
    \label{fig:pretrain_mixture}
\end{figure}

Next, we explore simple data mixture strategies. Our English dataset ratio, which is 50\%, is inspired by previous studies, including Typhoon~\citep{pipatanakul2023typhoon} and SambaLingo~\citep{csaki2023sambaefficientlyadaptingpretrainedlanguage} to mitigate catastrophic forgetting. For each large corpus, we repeat the data for 1 time, for smaller corpus (such as education documents and Wikipedia) we repeat the data up to three times. Additionally, we experimented with Doremi \citep{doremi} but did not observe a significant improvement. Ultimately, our best approach is based on a simple data mixture strategy where our English subset is sourced from \citet{slimpajama,redpajama}, and each of the datasets is repeated 1 time, except for the education content (2 times) and Wikipedia (3 times). The Thai subset of our pretraining data mixture is illustrated in Figure~\ref{fig:pretrain_mixture}.

\newpage
\subsection{Training} 
Based on our experiments, although current foundation models were already trained on some Thai texts, we do not extend the Thai tokenizer as done in Typhoon~\citep{pipatanakul2023typhoon}. This decision is because recent findings showed that adding more tokens to the tokenizer (which requires training their corresponding embeddings) can degrade overall performance \citep{sailor,zhao2024llamaenglishempiricalstudy} despite yielding high generation efficiency. We use this configuration during the model pre-training.

For all models, the AdamW optimizer is used in conjunction with a cosine learning rate scheduler. Gradient clipping is applied with a threshold value of 1.0. Our training is conducted using a context length of 8192. The DeepSpeed ZeRO optimization framework at Stage 2 without offloading is employed, as it provides the highest throughput in our experimental setup. All models are pre-trained on a single node comprising eight H100 GPUs. The learning rate is optimized individually for each model. We select the Llama~3 series \citep{llama3} and Qwen 2.5 series \citep{qwen2} as the base models due to their high performance on global leaderboards at the time of conducting pre-training. We perform full fine-tuning for models with 8B parameters or fewer and we apply LoRA \citep{hu2022lora} with a rank of 32 for models with 70B parameters. Due to our resource constraints, we train on sub-sampling instead of the full dataset.

\subsubsection*{Base Model}
Initially, we examined the performance of several 7–8B base models, as this is a standard size in academic scaling due to our resource constraints. Ultimately, we identified two families of base models that demonstrated significant potential for practical use:

1. \textbf{Llama 3 series}: A state-of-the-art open-source model developed by Meta. It is primarily trained for English performance. The instruction-tuned version also demonstrates excellent performance and is competitive with proprietary models.

2. \textbf{Qwen2.5 series}: A state-of-the-art open-source model developed by Alibaba. Its performance on knowledge-driven leaderboards is exceptional, surpassing many open-source and proprietary LLMs. It is optimized for English and Chinese as its primary languages.

As our Typhoon adaptation recipe is model-agnostic, we select both models as the base for the 7-8B category. Subsequently, we apply this recipe to other model sizes in the Llama 3 family since we observed a lower hallucination rate and better code-switching performance in our investigation.

\subsection{Evaluation} 
We evaluate the models using the ThaiExam and M3Exam datasets. While this evaluation method has been widely used for pre-trained language models, the scores obtained using these datasets are highly above the average level of a typical Thai person\footnote{\url{https://www.niets.or.th/th/content/view/11821}}. This can be attributed to contamination and saturation due to overfitting \citep{open-llm-leaderboard-v2}. Nevertheless, we utilize this signal to measure whether the model has acquired knowledge of the Thai language and context, as well as a development signal for improvement.

\subsection{Results and Findings}

\begin{table}[!ht]
  \tabcolsep=0.4mm
  \fontsize{8}{8}\selectfont
  % \small
  \begin{tabular}{lccccccccccc}
    \toprule
    \textbf{Model}           & \textbf{ThaiExam} & \textbf{ONET}  & \textbf{IC}    & \textbf{A-Level} & \textbf{TGAT}  & \textbf{TPAT}  & \textbf{M3Exam} & \textbf{Math}  & \textbf{Science} & \textbf{Social} & \textbf{Thai}  \\
    \midrule
    Llama3.2-1B              & 25.38             & 18.51          & \textbf{20.00} & \textbf{26.77}   & 32.30          & 29.31          & 25.30           & \textbf{23.52} & 25.36            & 27.48           & \textbf{24.82} \\
    Typhoon2-Llama-1B-base   & \textbf{26.83}    & \textbf{19.75} & 16.84          & 17.32            & \textbf{49.23} & \textbf{31.03} & \textbf{26.10}  & 21.71          & \textbf{25.60}   & \textbf{32.83}  & 24.27          \\
    \midrule
    Llama3.2-3B              & 40.42             & 30.86          & \textbf{46.31} & 20.47            & 63.07          & 41.37          & 36.81           & 21.71          & 36.23            & 50.74           & 38.54          \\
    Typhoon2-Llama-3B-base &  \textbf{44.53}& \textbf{40.12}& 40.00&  \textbf{26.77}&
    \textbf{69.23}& \textbf{46.55}& \textbf{41.84}& \textbf{24.43}& \textbf{41.30}& \textbf{60.07}& \textbf{41.56}\\
    \midrule
    Llama3.1-8B              & 45.80             & 38.27          & 46.31          & 34.64            & 61.53          & 48.27          & 43.33           & 27.14          & 40.82            & 58.33           & 47.05          \\
    Typhoon1.5-8B-base       & 48.82             & 41.35          & 41.05          & 40.94            & \textbf{70.76} & \textbf{50.00} & 43.88           & 22.62          & 43.47            & 62.81           & 46.63          \\
    Typhoon2-Llama-8B-base   & \textbf{51.20}    & \textbf{49.38} & \textbf{47.36} & \textbf{43.30}   & 67.69          & 48.27          & \textbf{47.52}  & \textbf{27.60} & \textbf{44.20}   & \textbf{68.90}  & \textbf{49.38} \\
    \midrule
    Qwen2.5-7B               & 55.74             & 51.23          & 60.00          & 41.73            & \textbf{72.30} & \textbf{53.44} & 55.65           & \textbf{46.15} & 54.10            & 66.54           & 55.82          \\
    Typhoon2-Qwen2.5-7B-base & \textbf{58.86}    & \textbf{58.64} & \textbf{65.26} & \textbf{55.11}   & 66.15          & 49.13          & \textbf{59.90}  & 42.98          & \textbf{59.42}   & \textbf{75.62}  & \textbf{61.59} \\
    \midrule

    Llama3.1-70B             & 60.74             & 62.34          & 67.36          & 53.54            & \textbf{66.15} & 54.31          & 60.35           & 38.91          & 62.56            & 76.99           & 62.96          \\
    Typhoon2-Llama-70B-base  & \textbf{63.38}    & \textbf{65.43} & \textbf{69.47} & \textbf{59.84}   & \textbf{66.15} & \textbf{56.03} & \textbf{62.33}  & \textbf{42.98} & \textbf{63.28}   & \textbf{78.60}  & \textbf{64.47} \\
    \midrule
    Avg. Human &- &31.80 &- &47.20 &40.60 &- &31.80 &- &- &- \\
    \bottomrule
  \end{tabular}
  \caption{The performance of pre-trained models on Exam in Thai. Human averages are estimated from \cite{onets_stats,tcas_stats,tpat1_stats}}
  \label{tab:pretrain_performance}
\end{table}

The evaluation results are shown in \Cref{tab:pretrain_performance}. We found the following insights from our experiments and evaluations.
\begin{itemize}
    \item \textbf{Each filtering subset has its own performance gain:} We observe a performance gain in the ``Science" score on ``High-Quality Text Selection," while we achieve a gain in the ``Social" score on ``Culture Related Text in Thai" and a gain in the ``Thai" score on the ``General subset" during the first-stage data mixture setup on M3Exam.
    \item \textbf{CPT-performance depends on the based model:} While CPT improves the model's performance, the based model's performance has a more significant effect on the overall score.
    \item \textbf{Data Mixture Effects in CPT Setup:} Qwen2.5 and Llama 3.1 respond differently to data mixtures, requiring distinct configurations to achieve comparable performance improvements. Exploring this phenomenon remains an area for future work.
    \item \textbf{Avoid Optimizing Solely for Knowledge:} While knowledge is a crucial aspect of LLMs, it is only one of many dimensions. Other objectives, such as instruction-following, task specificity, reasoning capabilities, and ease of parameter tuning, are equally vital for maximizing the overall utility of LLMs.
\end{itemize}

%% file: 3_posttraining.tex
\section{Post-training}

In our post-training process, the goal is to improve Typhoon's usability. We focus on instruction-following abilities, such as multi-turn response capabilities, system prompt following, and reasoning. We also focus on enhancing Typhoon 2's capabilities on tasks such as function calling and long context for both Thai and English. This section details our approaches and findings in the post-training stage of Typhoon 2.

\subsection{General Supervised Fine Tuning (SFT)}
\label{sec:general_dataset}
To make Typhoon 2 follow human instruction, we employ SFT as the principal strategy for aligning its outputs with human requirements. However, achieving effective human alignment is inherently multidimensional, as language models must accommodate a range of human values, preferences, and constraints. Recognizing this complexity, we design a comprehensive dataset encompassing multiple facets and specialized skills, enabling Typhoon 2 to meet diverse human needs better.

\subsubsection{Data}
We develop a dataset and combine it with the open-source dataset to ensure Typhoon 2's instruction-following performance based on these categories.

$\sbullet$ \textbf{English Instruction-Dataset}:
We combine several public datasets based on multiple iterations of our Typhoon development. In this version, we incorporate SystemChat\footnote{\url{https://huggingface.co/datasets/abacusai/SystemChat-1.1}}, Capybara\footnote{\url{https://huggingface.co/datasets/LDJnr/Capybara}}, OpenChat \citep{openchat} and a subset of the Tulu 3 \citep{tulu3} dataset as examples to retain the base model's English language comprehension. These datasets are selected primarily to align with human feedback patterns and various use cases in leveraging LLMs. For example, SystemChat supports system role-following features; Capybara facilitates multi-turn conversations, and OpenChat and Tulu 3 provide a rich diversity of instructions.

$\sbullet$ \textbf{Thai General Instruction-Dataset}:
We utilize the Typhoon self-instruct instruction dataset for training to align it to Thai. Unlike the original Typhoon \citep{pipatanakul2023typhoon} development, we do not use any translated English instruction data as it introduces hallucination.

$\sbullet$ \textbf{TyphoonIF Dataset}:
A new dataset is constructed based on AutoIF \citep{autoif}, available in both Thai and English. Seed constraints for the English and Thai versions are manually crafted to represent typical prompts that humans might use when interacting with LLMs in specific tasks. Additionally, random queries are taken from \texttt{airesearch/WangchanThaiInstruct} \citep{wangchaninstruct} and \texttt{Suraponn/thai\_instruction\_sft}\footnote{\url{https://huggingface.co/datasets/Suraponn/thai_instruction_sft}}, along with randomly sampled queries from the English dataset as previously described in this subsection. Rejected samples, identified through an instruction evaluation function, are filtered and added to the final dataset. The final instruction set comprises 150,000 question/instruction and response pairs. These pairs are randomly designated as either system or user turns to enhance generalization. To further improve cross-lingual transfer capabilities, instruction turns are randomly translated between Thai and English encouraging the models to align both Thai and English spaces to be similar to each other.

$\sbullet$ \textbf{Typhoon Personality Dataset}  
In addition, we curate an introductory prompt for Typhoon to incorporate its personality into the model. 

\begin{table}[h!]
\centering
\begin{tabular}{llrr}
  \toprule
  \small
  \textbf{Dataset} & \textbf{Subset} & \textbf{\# Examples} & \textbf{\# Tokens} \\
  \midrule
  \multirow{5}{*}{English Instruction-Dataset} 
  & SystemChat & 7 K & 4 M  \\
  & Capybara & 16 K & 15 M  \\ 
  & OpenChat & 10 K & 19 M  \\
  & Flan-fewshot & 50 K & 25 M \\
  & Others & 160 K & 88 M \\
  \midrule
  Thai General Instruction-Dataset & - & 10 K & 4 M \\
  TyphoonIF Dataset & - & 150 K & 49 M \\
  Typhoon Personality Dataset & - & 350 & 0.1 K \\
  \bottomrule
\end{tabular}
\caption{Detailed data composition of our general instruction-tuning.}
\label{tab:sft_mixture}
\end{table}

We present our SFT data mixture in Table~\ref{tab:sft_mixture}. The table shows the composition of the full general SFT dataset, highlighting its structure and the proportion of each component included.

\subsubsection{Experimental Setup}\label{sec:sft:exp}

\textbf{Training:} We perform full fine-tuning for SFT post-training, using the AdamW optimizer with a learning rate of 2e-5 for all experiments on the 7-8B model. We use a batch size of 16, and packing for a 32K context length during SFT. The SFT training is conducted for around 1,000 steps, resulting in approximately 700M tokens processed in total.

\textbf{Evaluation:} We evaluate the performance of general SFT using three datasets, focusing on assessing general instruction-following performance and the usability of LLMs in Thai and English. The three datasets are as follows:

\begin{itemize}
    \item \textbf{IFEval:} We employ IFEval \citep{ifeval}, a method designed to evaluate instruction-following capabilities using a set of verifiable instructions. These instructions are assessed against predefined rules implemented through test cases. The evaluation metric for IFEval is accuracy, which measures how well LLMs adhere to user-provided instructions. In addition to the standard IFEval (English version), we introduce \textbf{IFEval-TH}, a Thai version of IFEval. The original English instructions are translated into Thai, followed by a manual verification and correction process to ensure accuracy and content consistency. In this case, we evaluate the average of all four metrics from the original work.
    \item \textbf{Code-Switching:} We observed that English monolingual and English-Chinese bilingual LLMs exhibit a high tendency to produce code-switching responses when prompted to respond in Thai. To quantify this behavior, we propose a simple \textbf{code-switching evaluation} designed to assess the model's propensity to output non-Thai characters when following Thai instructions. The evaluation involves scenarios where the input instructions are sampled from the test subset of \texttt{airesearch/WangchanThaiInstruct}~\citep{wangchaninstruct}. The assessment is conducted across two temperature settings, $T=0.7$ and $T=1.0$, to measure the model's consistency in producing Thai-majority responses. 
    \item \textbf{MT-Bench}: We utilize MT-Bench, a variant of the LLM-as-a-judge evaluation framework, which employs a strong LLM to assess responses to open-ended questions based on correctness, fluency, and adherence to instructions. For the \textbf{ThaiLLM leaderboard}~\citep{thaillm-leaderboard}, we use the Thai version of MT-Bench developed by VISTEC \citep{mtbenchth}, while the English version follows the LMSYS implementation~\citep{mtbench}.
\end{itemize}

For Code-switching (CS) evaluation, the metric is \textbf{accuracy}, defined as: 
\begin{enumerate}
    \item The response does not contain characters from other languages.
    \item Thai characters constitute the majority of the response content.
\end{enumerate}

\textbf{Baseline}: We compare the Typhoon 2 model based on Llama 8B fine-tuning on the newly developed dataset with Typhoon 1.5 (Llama-based) and Llama 3.1 8B Instruct.

\subsubsection{Results and Findings}
We present our results and findings from the selection process of our SFT dataset in \Cref{tab:general_sft}.

\begin{table}[htbp]
    \centering
    % \fontsize{8}{9}\selectfont
    % \footnotesize
    % \tabcolsep=0.3mm
    \begin{tabular}{rcccccccc}
        \toprule
         \multirow{2}{*}{\textbf{Model}}& \multicolumn{2}{c}{\textbf{IFEval}} & \multicolumn{2}{c}{\textbf{MT-Bench}} & \multicolumn{2}{c}{\textbf{Code-switch}} \\
         &\textbf{TH} &\textbf{EN} &\textbf{TH} &\textbf{EN} &\textbf{1.0} &\textbf{0.7} \\
         \midrule
Llama3.1-8B-Instruct&	58.04&	\textbf{77.64}&	5.11&	\textbf{8.12}&	11.20&	93.00\\
Typhoon1.5-8B-Instruct&	58.68&	71.33&	5.18&	7.34&	\textbf{98.80}&	98.60\\
Typhoon2-Llama-8B-Instruct& \textbf{72.60}& 76.43& \textbf{5.74}& 7.58& 98.00& \textbf{98.80}\\
 \bottomrule
    \end{tabular}
    \caption{Performance on General instruction following}
    \label{tab:general_sft}
\end{table}

We found the following insights:

\textbf{Impact of English Data on Thai Performance}:
Our preliminary experiments indicate that the inclusion of a high-quality English dataset contributes to improving the performance of Thai language models. This suggests that cross-lingual benefits can be obtained when using high-quality data from a related or dominant language.

\textbf{Thai-English Ratio}:
To enable the model to generate fluent responses in Thai, we initially experimented with a Thai-English ratio of 1:9. Despite the low proportion of Thai data, the model demonstrated the ability to respond adequately in Thai. However, increasing the ratio of Thai data leads to a performance improvement. After conducting multiple trials, we empirically found that the optimal Thai-to-English ratio is \textbf{3:7}.

\textbf{Importance of Data Quality}: The presence of low-quality data from a single source can lead to a performance degradation of up to 20\% across the entire system.

\textbf{Quality over Quantity}: We examine combining multiple large datasets, but the performance levels are only comparable to or even worse than those achieved with a smaller curated dataset tailored to specific functions and use cases.

\subsection{Domain-Specific SFT}
The model's performance on general domain instruction-following tasks is satisfactory across all datasets. However, a noticeable drop in domain-specific abilities, particularly in coding and mathematics, is observed. To address this issue, we examine incorporating math and coding instruction data in the experiment.

\subsubsection{Data}

\noindent \textbf{Math \& Code Dataset}

To improve math and coding performance, we examine domain-specific datasets and select three based on their competitive results as follows,

\begin{itemize}
    \item \textbf{Dart-math} \citep{dartmath}: An augmented math dataset verified using a rejection sampling method, focusing on complex questions. The answers are sampled from DeepSeekMath \citep{deepseek-math}.
    \item \textbf{ScaleQuest-Math} \citep{scalequest}: A technique to scale diverse math instruction using a small seed math problem. The responses are sampled through a combination of DeepSeekMath-RL \citep{deepseek-math} and Qwen2.5-Math \citep{yang2024qwen25mathtechnicalreportmathematical}.
    \item \textbf{OpenCoder-Instruct} \citep{opencoder}: Stage 2 instruction dataset from the OpenCoder project contributes to the strong performance of fully open code LLMs. The dataset is created by leverage multiple methods to scale code instruction dataset such as Magicoder \citep{wei2024magicoderempoweringcodegeneration} and Wizardcoder \citep{luo2023wizardcoderempoweringcodelarge}.
\end{itemize}

We also translate a subset of each dataset into Thai using an early version of our Typhoon2 model. Invalid translations are filtered out by validating the responses against the final answers of the math solutions, and we use an LLM as a judge to evaluate coding correctness. In total, we translate approximately 12K examples in the dataset.

\textbf{Data Mixture}

The dataset consists of approximately 250,000 code samples and 200,000 math problems. Code samples are sourced from \texttt{OpenCoder-Instruct}, while math problems are collected from two sources, \texttt{ScaleQuest-Math} and \texttt{Dart-Math}, in an equal proportion (1:1). Preliminary experiments indicated that combining both math datasets results in better performance compared to using a single source. Additionally, a Thai-translated subset is incorporated, containing 6,000 samples per domain (code and math).

\subsubsection{Experimental Setup}
We evaluate the Typhoon 2 models' performance under various training scenarios, focusing on the impact of code and math domain-specific subsets. Three key evaluations are conducted as follows,
\begin{itemize}
    \item  \textbf{Training Data Impact}: Models trained on the \textit{General-only} subset are compared to those using \textit{General + Code \& Math} to assess the benefits of domain-specific data.
    \item \textbf{Math Performance}: Typhoon 2, fine-tuned on \textit{General + Code \& Math}, is compared to Llama 3.1 8B Instruct and Qwen2.5 7B Instruct to evaluate math task effectiveness.
    \item \textbf{Code Performance}: Typhoon 2's code performance, also fine-tuned on \textit{General + Code \& Math}, is compared against the same base models to assess the coding ability.
\end{itemize}

\textbf{Training}: The training process and hyperparameters are identical to those described in~\Cref{sec:sft:exp} for General-SFT training.

\textbf{Evaluation Data}: We evaluate hard domain-specific tasks for LLMs, such as coding and math, which are also related to reasoning performance. We also incorporate evaluations from the general domain to ensure the model does not overfit specific domain tasks.

\begin{itemize}
    \item \textbf{GSM8K}: The Grade School Math 8K \citep{gsm8k} consists of diverse grade school math word problems which are basic mathematical problems that require multiple-step reasoning.
    \item \textbf{MATH}: The Mathematics Aptitude Test of Heuristics \citep{math} is a hard math dataset consisting of problems from mathematics competitions.
    \item \textbf{HumanEval}: HumanEval \citep{humaneval} consists of programming problems with a function signature, docstring, body, and several unit tests.
    \item \textbf{MBPP}: MBPP \citep{mbpp} is a set of Python programming problems designed to be solvable by entry-level programmers, covering programming fundamentals, standard library functionality, and related topics.
\end{itemize}

\textbf{Evaluation Implementation:} The mathematical evaluation is zero-shot based on the \texttt{dartmath} implementation \citep{dartmath}. Code evaluation is zero-shot based on the \texttt{evalplus} implementation \citep{evalplus} -- we report the base subset result. The Thai subset is created by directly translating the original dataset into Thai using GPT-4o, with automatic verification performed using the LLM-as-judge technique.

\subsubsection{Results and Findings}

\begin{table}[htbp]
    \centering
    % \small
    \tabcolsep=0.5mm
    \begin{tabular}{rcccccccc}
    \toprule
         \textbf{Model}& \textbf{IFEval(Avg)}& \textbf{MT-Bench(Avg)}& \textbf{GSM8K} & \textbf{Math} & \textbf{HumanEval} & \textbf{MBPP} \\
         \midrule
 General only& 73.75& 6.581& 39.10& 16.34& 49.70&54.35\\
 General + Code \& Math& \textbf{74.7}& \textbf{6.715}& \textbf{81.00}& \textbf{49.04}& \textbf{63.70}& \textbf{61.90}\\
 \bottomrule
    \end{tabular}
    \caption{Typhoon2-Llama-8B Performance difference when adding domain-specific dataset}
    \label{tab:math_code}
\end{table}

\begin{table}[htbp]
    \centering
    % \small
    % \tabcolsep=0.8mm
    \begin{tabular}{rcccccccc}
    \toprule
         \textbf{Model}&  \textbf{GSM8K-TH} & \textbf{GSM8K-EN} & \textbf{Math-TH} & \textbf{Math-EN}\\
\midrule
Llama3.1-8B-Instruct&	45.18&	62.40&	24.42&	48.00\\
Typhoon2-Llama3.1-8B-Instruct&   \textbf{71.72}&	\textbf{81.00}&	\textbf{38.48}&	\textbf{49.04}\\
\midrule
Qwen2.5-7B-Instruct&47.53&	81.00&	17.41&	\textbf{73.40}\\         
Typhoon2-Qwen2.5-7B-Instruct& \textbf{79.07}&	\textbf{84.20}&	\textbf{55.42}&	66.42\\
 \bottomrule
    \end{tabular}
    \caption{Math only performance compared to SOTA model}
    \label{tab:math}
\end{table}

\begin{table}[htbp]
    \centering
    % \small
    \tabcolsep=0.6mm
    \begin{tabular}{rcccccccc}
    \toprule
         \textbf{Model}& \textbf{HumanEval-TH} & \textbf{HumanEval-EN} & \textbf{MBPP-TH} & \textbf{MBPP-EN} \\
         \midrule
Llama3.1-8B-Instruct& 51.8&	67.7&	\textbf{64.6}&	\textbf{66.9}\\
Typhoon2-Llama3.1-8B-Instruct& \textbf{58.5}&	\textbf{68.9}&	60.8&	63.0\\
\midrule
Qwen2.5-7B-Instruct& 58.5&	68.9&	60.8&	63.0\\
Typhoon2-Qwen2.5-7B-Instruct& \textbf{73.2}&	\textbf{79.3}&	\textbf{78.3}&	\textbf{81.7}\\

 \bottomrule
    \end{tabular}
    \caption{Code only performance compare to SOTA model}
    \label{tab:code}
\end{table}

Evaluation results for math and code subsets are presented in \Cref{tab:math} and \Cref{tab:code}, respectively. We found the following insights:

\begin{itemize}
    \item \textbf{Math \& Code also improve general performance}: The addition of the Math \& Code data shows an improvement to the overall performance, as indicated by the IFEval/MT-Bench results presented in Table \ref{tab:math_code}.
    \item \textbf{Math performance in English does not automatically transfer to Thai}: Results in Tables \ref{tab:math} and \ref{tab:code} highlight that state-of-the-art LLMs, while exhibiting strong mathematical capabilities in English, show a notable drop in performance when applied to Thai. In contrast, performance on coding tasks remains similar between English and Thai.
    \item \textbf{Larger Model Requires Fewer Data Points}: In our adaptation study with a 70B model, the model reaches its performance saturation using only 50K coding samples and 60K math problems. In comparison, the smaller 7B model requires the entire dataset, consisting of 250K coding samples and 200K math problems, to achieve a similar performance level. Based on these findings, we use 50K coding samples and 60K math problems for the final evaluation of the 70B model.
\end{itemize}

\subsection{Long Context}
% TODO: Add references
% e.g., Longalign, anti-haystack, iapp wiki qa
The capability to handle long contexts is essential for LLMs to process and understand complex and lengthy texts. Numerous real-world applications, such as summarizing academic papers and analyzing legal documents, demand models capable of handling inputs that go beyond standard context length limitations. However, training LLMs with extended context sizes is computationally demanding, often requiring significant training time and GPU resources. In practice, this limitation typically restricts the maximum context length to 32,768 tokens on four A100 (80GB) GPUs with Deepspeed ZeRO Stage 3. To enhance the long-context capabilities of Typhoon 2, we extend its context length from 8,192 tokens in Typhoon 1.5 to 32,768 tokens and generalize support for context lengths up to 128K tokens. We evaluate this capability through experiments on SFT tasks following CPT with an 8,192-token context size. 

\subsubsection{Data}
We construct the dataset based on \Cref{sec:general_dataset} combined with three primary sources to ensure both effective exploitation of long contexts and robust multilingual support. These three primary sources are:

\begin{itemize}
    \item \textbf{LongAlign} \citep{bai-etal-2024-longalign}: This dataset is derived from the LongAlign framework, which is specifically designed to enhance LLMs for long-context understanding and instruction-following. While the full framework encompasses data creation, training strategies, and evaluation for long-context alignment, we selectively incorporate the dataset component for our work.
    \item \textbf{Anti-haystack\footnote{\url{https://huggingface.co/datasets/wenbopan/anti-haystack}}}: This dataset is designed for enhancing the ability of LLMs to locate short and precise facts within long, noisy documents, emulating a ``needle in a haystack" challenge.
    \item \textbf{IApp-Wiki-QA Dataset} \citep{kobkrit_viriyayudhakorn_2021_4539916}: To enhance Thai long-context capabilities, we utilize the Thai Wikipedia Question Answering dataset, consisting of 1,500 unique context records. We extend and reformat the data using the following processes:
    \begin{itemize}
        \item \textbf{Irrelevant Content Addition:} Random irrelevant contents are sampled from the ThaiSum dataset \citep{chumpolsathien_2020}, a news summarization dataset in the Thai language, to introduce noise and increase the context length.
        \item \textbf{Question and Answer Integration:} Questions and answers from the IApp-Wiki-QA dataset are randomly positioned within the extended context.
        \item \textbf{Token Length Requirement:} Each row in the dataset is structured to ensure that the Thai token count exceeds 30,000 tokens.
    \end{itemize}
\end{itemize}

\subsubsection{Experimental Setup}

\noindent \textbf{Evaluation}:  
We evaluate the long context abilities using the Needle-in-a-Haystack (NIAH) method~\citep{kamradt2023needle}. This framework assesses a model's ability to retrieve specific ``needle sentences" hidden within random segments of lengthy documents.  

The evaluation involves completing sentences such as:  
\begin{quote}
    \textit{``The best thing to do in San Francisco is to eat a sandwich and sit in Dolores Park on a sunny day."}  
\end{quote}
where the corresponding input prompt is:  
\begin{quote}
    \textit{``What is the best thing to do in San Francisco?"}  
\end{quote}

We evaluate the long context abilities in both English and Thai. To ensure linguistic consistency, the Thai dataset was created using our in-house machine translation model to translate the English dataset.

% \noindent \textbf{Experiment Design}  
This work considers two model architectures:  

\begin{itemize}
    \item \textbf{Llama3.1-based Model}~\citep{llama3}: The model underwent continued pre-training with a context length of 8,192 tokens, using the original RoPE~\citep{su2023roformerenhancedtransformerrotary} base frequency hyperparameter of 500,000. This was followed by supervised fine-tuning to extend the context length to 32,768 tokens.  

    \item \textbf{Qwen2.5-based Model}~\citep{qwen2}: A similar training pipeline was applied, starting with pre-training using Qwen2.5's original RoPE~\citep{su2023roformerenhancedtransformerrotary} configuration, which employs a base frequency of 1,000,000. Additionally, this model leverages the YARN mechanism~\citep{peng2024yarn}, optimized for long-context scenarios.  
\end{itemize}

\newpage
\subsubsection{Results and Findings}

\noindent \textbf{Findings from Typhoon2-Llama3.1-8B,70B-Instruct Evaluation}  
\begin{figure}[H]
    \centerline{
        \includegraphics[width=\linewidth,keepaspectratio]{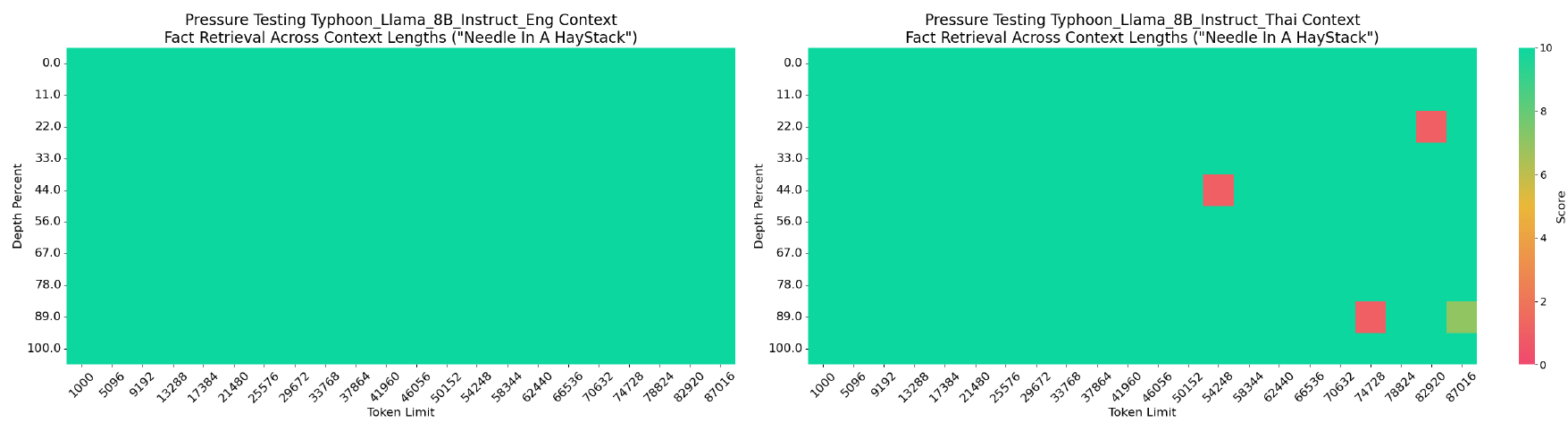}}
    \caption{Evaluation of Typhoon2-Llama3.1-8B-Instruct on Needle-in-a-Haystack for both English (Left) and Thai (Right).}
    \label{fig:typhoon_llama_8b_long}
\end{figure}

\begin{figure}[H]
    \centerline{
        \includegraphics[width=\linewidth,keepaspectratio]{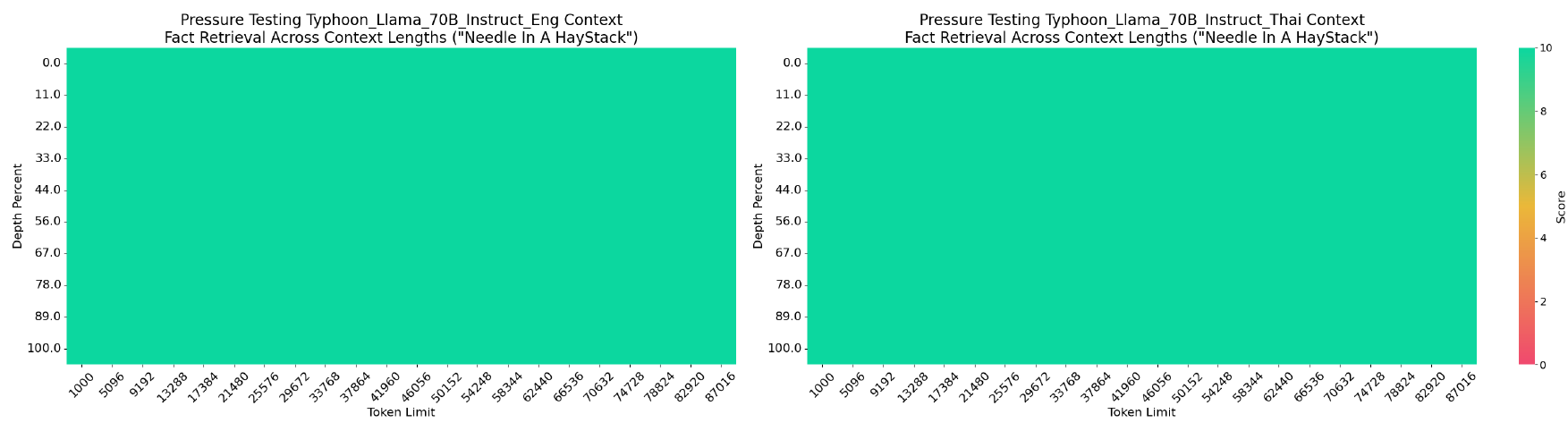}}
    \caption{Evaluation of Typhoon2-Llama3.1-70B-Instruct on Needle-in-a-Haystack for both English (Left) and Thai (Right).}
    \label{fig:typhoon_llama_70b_long}
\end{figure}

As illustrated in Figures~\ref{fig:typhoon_llama_8b_long} and~\ref{fig:typhoon_llama_70b_long}, both Typhoon2-Llama-3.1-8B,70B-Instruct models support a maximum context length of approximately 90,000 tokens. This is a reduction compared to the original Llama 3.1 model, which supports up to 128,000 tokens. We hypothesize that this limitation is due to two key factors:  

\begin{enumerate}
    \item The original Llama 3.1 model was trained \textbf{incrementally across multiple stages}, progressively extending its context length to 128,000 tokens.  
    \item Our CPT approach is restricted to a context length of 8,192 tokens, potentially limiting the model's ability to generalize to longer contexts, despite adjustments to the RoPE scaling \citep{su2023roformerenhancedtransformerrotary}.  
\end{enumerate}

Addressing these limitations could pave the way for future enhancements to the Llama-based Typhoon 2 models.

\newpage
\noindent \textbf{Findings from Typhoon2-Qwen2.5-7B-Instruct Evaluation}  
\begin{figure}[ht]
    \centerline{
        \includegraphics[width=\linewidth,keepaspectratio]{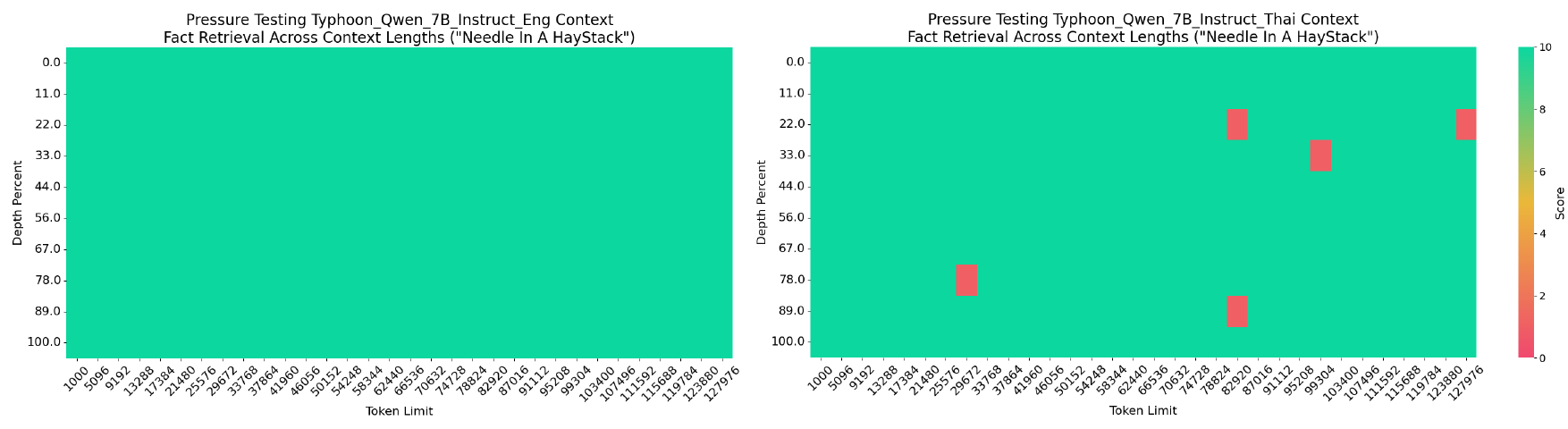}}
    \caption{Evaluation of Typhoon2-Qwen2.5-7B-Instruct on Needle-in-a-Haystack for both English (Left) and Thai (Right).}
    \label{fig:typhoon_qwen_8b_long}
\end{figure}

As shown in Figure~\ref{fig:typhoon_qwen_8b_long}, Typhoon2-Qwen2.5-7B-Instruct successfully supports a maximum context length of 128,000 tokens, matching the original performance of Qwen2.5. Remarkably, despite training with shorter context lengths, the Qwen-based Typhoon model demonstrates effective extrapolation to significantly longer contexts, surpassing the 32,768-token range.

\noindent \textbf{Dataset Composition and Recommendations:}  
Our analysis highlights the importance of data composition in achieving robust long-context performance, as follows:  

1. Models trained exclusively on English long-context tasks often \textbf{underperform} on Thai tasks, but incorporating Thai long-context data into training can mitigate this problem.  

2. Training solely on short-context data leads to substantial degradation in long-context performance. Conversely, overemphasis on long-context data negatively impacts performance on other benchmarks.  

Based on our experiments, a good dataset composition consists of 15\% long-context data (with a 20:80 ratio of Thai to English) and 85\% short-context data. This approach yields the best overall performance across both long- and short-context tasks.

\subsection{Function Calling}
Function calling is essential for enhancing the capabilities of LLMs to tackle complex, real-world tasks. By enabling LLMs to interact with external tools and third-party services, function calling facilitates impactful applications such as workflow automation and financial analysis. To equip the Typhoon models with this capability, we utilize existing function-calling datasets described in the following subsection.

\subsubsection{Data}
We construct the general instruction dataset from Section~\ref{sec:general_dataset} combined with three sources to ensure diversity, quality, and multilingual support:

\begin{itemize}
    \item \textbf{APIGen}~\citep{liu2024apigen}: This synthetic dataset emphasizes diversity and quality through a multi-stage hierarchical verification process. A Flan-style approach is used to generate outputs in various formats (e.g., JSON, YAML, XML). An in-house translation system is employed to translate data from English to Thai, ensuring robust bilingual representation.
    \item \textbf{ToolACE}~\citep{liu2024toolacewinningpointsllm}: This dataset builds on a prior study, focusing on enhancing the function-calling capabilities of LLMs. Synthetic data is translated into Thai using the same in-house machine translation system, enabling bilingual support and increasing dataset complexity and diversity.
    \item \textbf{Glaive-v2\footnote{\url{https://huggingface.co/datasets/glaiveai/glaive-function-calling-v2}}}: The popular \texttt{Glaive-function-calling-v2} dataset is incorporated into the training data mix, complementing APIGen and ToolACE datasets to provide a comprehensive foundation for training.
\end{itemize}

This data combination ensures high-quality, diverse, and multilingual data to train and evaluate models effectively.

\subsubsection{Experimental Setup}

\noindent \textbf{Evaluation}: The trained models are evaluated using the Berkeley Function-Calling Benchmark (BFCL)~\citep{patil2023gorilla}, a comprehensive framework for assessing the function-calling capabilities of large language models across diverse domains. The evaluation is conducted in both English and Thai, with the Thai dataset generated using our in-house machine translation from the English dataset. The assessment consists of two key components:

\begin{itemize}
    \item \textbf{Abstract Syntax Tree (AST) Evaluation}: This component measures the syntactic correctness of generated function calls by comparing them to predefined specifications, focusing on function names, parameters, and data types.
    \item \textbf{Executable Function (Exec) Evaluation}: This component verifies operational correctness by executing the generated functions to ensure they compile and perform as intended.
\end{itemize}

\noindent \textbf{Baselines}: For evaluation, we compare our model, Typhoon 2, against open-weight models, including variants of Qwen2.5 and Llama 3.1, to benchmark performance.

\subsubsection{Results and Findings}

\begin{table}[!ht]
    \centering
    \footnotesize
    \begin{tabular}{lcccccccc}
    \toprule
     \textbf{Model} & \textbf{Overall}  & \textbf{AST} & \textbf{Exec} & \textbf{Live} & \textbf{MultiTurn} & \textbf{Relv} & \textbf{Irrelv} \\
    \midrule
    \rowcolor{Gray}
        \multicolumn{8}{l}{\textbf{1B}} \\
    Typhoon2-Llama-1B-Instruct  & \textbf{45.60} & \textbf{64.15} & \textbf{66.20} & \textbf{49.53} & \textbf{24.00} & \textbf{82.93} & 52.99  \\
    Qwen2.5-1.5B-Instruct          & 36.50 & 59.40 & 57.96 & 39.14 & 16.62 & 73.17 & 24.05 \\
    Llama3.2-1B-Instruct        & 17.88 & 21.62 & 19.73 & 29.99 & 0.12 & 46.34 & \textbf{53.75} \\
    \rowcolor{Gray}
        \multicolumn{8}{l}{\textbf{3B}} \\
    Typhoon2-Llama-3B-Instruct  & \textbf{75.90} & 77.85 & 78.77 & \textbf{66.50} & \textbf{81.25} & 74.61 & \textbf{83.20}  \\
    Qwen2.5-3B-Instruct          & 62.55 & 80.56 & 75.27 & 60.06 & 51.38 & \textbf{87.80} & 55.08 \\
    Llama3.2-3B-Instruct        & 53.87 & \textbf{81.17} & \textbf{80.48} & 55.44 & 27.00 & 82.93 & 55.43 \\
    \rowcolor{Gray}
        \multicolumn{8}{l}{\textbf{7-8B}} \\
    Typhoon2-Qwen-7B-Instruct  & \textbf{79.08} & 83.21 & 83.05 & 68.90 & \textbf{84.50} & \textbf{92.68} & 77.88 \\
    Typhoon2-Llama-8B-Instruct & 75.44 & 81.88 & 79.00 & \textbf{70.15} & 74.25 & 85.37 & \textbf{84.19} \\
    Qwen2.5-7B-Instruct          & 74.81 & 83.92 & \textbf{85.00} & 65.30 & 75.75 & 85.37 & 65.23 \\
    Openthaigpt1.5-7B-Instruct   & 73.10 &  82.98 & 81.64 & 64.37 & 73.62 & 87.80 & 64.04 \\
    Llama3.1-8B-Instruct        & 65.09 & \textbf{83.98} & 83.36 & 57.71 & 58.00 & 78.05 & 41.73 \\
    \rowcolor{Gray}
        \multicolumn{8}{l}{\textbf{70B}} \\
    Typhoon2-Llama-70B-Instruct  & \textbf{65.78} & \textbf{90.29} & 83.36 & \textbf{76.63} & \textbf{33.62} & 75.61 & \textbf{80.35}  \\
    Llama-3.3-70B-Instruct  & 56.36 & 87.10 & \textbf{86.89} & 65.57 & 18.25 & \textbf{97.56} & 58.88  \\
    Llama-3.1-70B-Instruct  & 53.61 & 88.73 & 83.68 & 61.00 & 15.00 & 92.68 & 57.56  \\
    \bottomrule
    \end{tabular}
    \caption{BFCL V3 Benchmark on \textit{English} Dataset where we report accuracies for overall, AST, Exec, Live, Multi-turn as well as relevance and irrelevance scores.}
    \label{tab:bfcl_en}
\end{table}

\begin{table}[!ht]
    \centering
    \footnotesize
    \begin{tabular}{lcccccccc}
    \toprule
     \textbf{Model} & \textbf{Overall}  & \textbf{AST} & \textbf{Exec} & \textbf{Live} & \textbf{MultiTurn} & \textbf{Relv} & \textbf{Irrelv} \\
    \midrule
    \rowcolor{Gray}
        \multicolumn{8}{l}{\textbf{1B}} \\
    Typhoon2-Llama-1B-Instruct  & \textbf{34.96}&	\textbf{45.31}&	\textbf{60.05}&	\textbf{36.12}&	18.75&	\textbf{92.68}&	32.26  \\
    Qwen2.5-1.5B-Instruct          & 29.88&	32.69&	50.98&	30.25&	\textbf{20.62}&	60.98&	25.54 \\
    Llama3.2-1B-Instruct        & 13.83&	12.98&	0.18&	26.61&	0.75&	41.46&	\textbf{67.01} \\
    \rowcolor{Gray}
        \multicolumn{8}{l}{\textbf{3B}} \\
    Typhoon2-Llama-3B-Instruct  & \textbf{71.36}&	\textbf{61.92}&	\textbf{73.48}&	\textbf{56.77}&	\textbf{87.50}&	78.05&	\textbf{74.70}  \\
    Qwen2.5-3B-Instruct          & 53.78&	55.71&	70.55&	48.73&	49.88&	\textbf{82.93}&	54.19 \\
    Llama3.2-3B-Instruct        & 35.43&	37.40&	59.89&	33.90&	25.00&	\textbf{82.93}&	34.92 \\
    \rowcolor{Gray}
        \multicolumn{8}{l}{\textbf{7-8B}} \\
    Typhoon2-Qwen-7B-Instruct  & \textbf{75.12} & \textbf{71.00} & \textbf{76.62} & 57.44 & \textbf{95.38} & \textbf{87.80} & 55.65  \\
    Typhoon2-Llama-8B-Instruct & 74.24 & 70.79 & 74.05 & \textbf{64.68} & 84.00 & \textbf{87.80} & \textbf{78.86} \\
    Qwen2.5-7B-Instruct          & 66.06 & 57.48 & 77.79 & 54.69 & 75.75 & 85.37 & 61.52 \\
    Openthaigpt1.5-7B-Instruct   & 65.53 & 59.77 & 75.37 & 53.35 & 75.62 & 80.49 & 60.47 \\
    Llama3.1-8B-Instruct        & 36.92 & 50.94 & 76.18 & 39.63 & 12.88 & 82.93 & 18.66 \\
    \rowcolor{Gray}
        \multicolumn{8}{l}{\textbf{70B}} \\
    Typhoon2-Llama-70B-Instruct  & \textbf{70.89}&	\textbf{78.83}&	\textbf{82.32}&	\textbf{67.88}&	\textbf{64.38}&	90.24&	\textbf{70.21}  \\
    Llama-3.3-70B-Instruct  & 50.30&	68.21&	80.71&	56.91&	21.75&	\textbf{97.56}&	49.86 \\
    Llama-3.1-70B-Instruct  & 48.79&	69.92&	81.95&	47.89&	26.88&	92.68&	33.00  \\
    \bottomrule
    \end{tabular}
    \caption{BFCL V3 Benchmark on \textit{Thai} Dataset where we report accuracies for overall, AST, Exec, Live, Multi-turn as well as relevance and irrelevance scores.}
    \label{tab:bfcl_th}
\end{table}

\noindent \textbf{Performance Enhancement in English and Thai}: Tables ~\ref{tab:bfcl_en} and ~\ref{tab:bfcl_th} demonstrate that our English-Thai data mixing approach leads to substantial performance improvements across both languages. Specifically, the Typhoon2-Qwen2.5-7B model consistently achieves the highest overall accuracy in both English (79.08\%) and Thai (75.12\%), outperforming other baselines, including OpenThaiGPT-1.5, which is also a Qwen2.5 based model. Notably, Qwen-based Typhoon models outperform their Llama-based counterparts across all evaluation metrics, emphasizing the effectiveness of Qwen in this multilingual setting.

\noindent \textbf{Data Proportion and Balance:} Our results indicate that the proportion of token data from tool-calling datasets should not exceed or equal that of instruction-following datasets. Maintaining an appropriate balance is critical to ensuring practical model training.

\noindent \textbf{Accuracy vs. Generalization:} While the model demonstrates high accuracy on tool-calling tasks, it exhibits limited generalization capabilities across other tasks. This highlights the need for a more diverse and balanced training dataset.

\noindent \textbf{Dataset Composition Recommendations:}
\begin{itemize}
    \item Tool-calling data should comprise \textbf{5-10\% of the total dataset} to balance specialization with generalization.
    \item Thai-translated tool-calling data should represent approximately \textbf{40\% of the tool-calling} subset to ensure adequate multilingual support.
    \item \textbf{High-quality instruction-following datasets} are essential for improving both generalization and tool-calling performance, suggesting a synergistic relationship between these components.
\end{itemize}

\subsection{Distillation}
Distillation is a widely recognized method to transfer knowledge from a stronger model to a smaller model, as seen from the era of BERT to DistilBERT\citep{distilbertdistilledversionbert}, and so on. In the case of LLMs, there have also been successful examples such as Minitron \citep{sreenivas2024llmpruningdistillationpractice} and Llama 3.2\footnote{\url{https://ai.meta.com/blog/llama-3-2-connect-2024-vision-edge-mobile-devices/}}. Following this approach, we apply a distillation technique to enhance the performance of the smaller Typhoon models.

\subsubsection{Top-k Logits Distillation}

We employ top-k logits distillation, drawing inspiration from Arcee-AI's methodology\footnote{\url{https://blog.arcee.ai/introducing-arcee-supernova-medius-a-14b-model-that-rivals-a-70b-2}}. The objective of this approach is to transfer knowledge from a larger teacher model to a smaller student model. Our method involves constructing logit data from a larger teacher model, where only the top-k predictions for each vocabulary token are retained. The top-k logits are obtained from early versions of the Llama-based Typhoon 2 models, specifically the 8B and 70B variants. This method assumes that the teacher and student models share identical vocabularies.

To achieve effective knowledge distillation, the loss function includes Kullback-Leibler (KL) divergence for logits alignment and a cross-entropy loss for supervised training. The formulation of the loss function is as follows:

\[
L_{\text{KD}} = \alpha \cdot T^2 \cdot \text{KL}\left(\sigma\left(\frac{\mathbf{z}_\text{student}^{(k)}}{T}\right) \parallel \sigma\left(\frac{\mathbf{z}_\text{teacher}^{(k)}}{T}\right)\right) + (1 - \alpha) \cdot L_{\text{CE}}(\mathbf{y}_\text{true}, \mathbf{z}_\text{student})\text{,}
\]

where \( L_{\text{KD}} \) denotes the knowledge distillation loss, \( \mathbf{z}_\text{student}^{(k)} \) denotes top-k logits from the student model, \( \mathbf{z}_\text{teacher}^{(k)} \) denotes top-k logits from the teacher model, \( \sigma \) denotes the softmax operation, \( T \) denotes a temperature hyperparameter, \( \alpha \) denotes a weight for balancing between the two loss components, \( L_{\text{CE}} \) is a cross-entropy loss of the ground-truth labels (\( \mathbf{y}_\text{true} \)), and \( k \) denotes the number of top logits retained.

In our configuration, the hyperparameters are set as follows: \( \alpha = 0.5 \), \( k = 8 \) and \( T = 1 \). This combination ensures a balanced trade-off between the distillation objective (matching the top-k logits of the teacher) and the supervised training objective (matching true labels). 

\subsubsection{Experimental Setup}
In this experiment, we compare two approaches: (1) SFT-only and (2) combining SFT with top-k distillation. Both approaches are experimented on three datasets: 1) English Instruction, 2) Thai General Instruction, and 3) TyphoonIF. These datasets are the same as those used in \Cref{sec:sft:exp}. The experiments utilize Llama-based Typhoon 1B as the base model. For all experiments, we apply the same learning rate used for SFT.

\textbf{Evaluation:} We evaluate the distillation technique following the same approach as in \Cref{sec:sft:exp}, which are used for evaluating instruction-following tasks.

\subsubsection{Results and Findings}
The results of the distillation experiment are presented in \Cref{tab:distillation}.

\begin{table}[htbp]
    \centering
    % \fontsize{8}{9}\selectfont
    % \tabcolsep=0.8mm
    \begin{tabular}{rcccccccc}
    \toprule
         \multirow{2}{*}{\textbf{Model}}& \multicolumn{2}{c}{\textbf{IFEval}} & \multicolumn{2}{c}{\textbf{MT-Bench}} & \multicolumn{2}{c}{\textbf{Code-switch}} \\
         &\textbf{TH} &\textbf{EN} &\textbf{TH} &\textbf{EN} &\textbf{1.0} &\textbf{0.7} \\
         \midrule
 SFT& 49.53& \textbf{53.76}& 3.68& 5.37& 85.40& 92.20\\
 Distillation& \textbf{52.46}& 53.35& \textbf{3.97}& \textbf{5.40}& \textbf{88.00}& \textbf{96.40}\\
 \bottomrule
    \end{tabular}
    \caption{Performance comparison between standard SFT and distillation}
    \label{tab:distillation}
\end{table}

We found the following key insights:
\begin{itemize}
    \item \textbf{Impact of Distillation on Performance}: Based on the results, distillation yields a performance improvement in most of the aspects of the small model.
    \item \textbf{Does a larger model's logits improve performance?} In our preliminary experiments, we distil the logits from both Typhoon2-8B and Typhoon2-70B models. Based on this setup, we observe a similar results on the 1B and 3B models in our settings.
\end{itemize}

\subsection{Model Merging}

Model merging, a method to combine the weights of two models, has recently been shown to improve performance in LLMs \citep{akiba2024evolutionaryoptimizationmodelmerging} . We previously performed model merging for our Typhoon 1.5X series\footnote{\url{https://blog.opentyphoon.ai/typhoon-1-5x-our-experiment-designed-for-application-use-cases-7b85d9e9845c}}, resulting in significant improvements for Thai and English instruction-following tasks. Notably, our larger-size models, such as those with 70B parameters, demonstrated remarkable performance gains. Therefore, we apply model merging techniques to our 70B models in this iteration.

\subsubsection{Experimental Setup}
In our experiment involving the Arcee-AI Mergekit \citep{mergekit}, we explore merging methods implemented in the Arcee-AI Mergekit, such as linear, slerp, TIES \citep{yadav2023tiesmerging}, and DARE \citep{yu2024languagemodelssupermario}. We manually search for each merge hyperparameter.

\textbf{Model to Merge}: Our strategy is to merge the model with the strongest performance in its family--based on the same pretraining foundation and selected according to its pre-trained model. In this case, we merge Llama-based Typhoon 2 70B SFT with Llama 3.3 70B Instruct.

\textbf{Evaluation}: We evaluate the merging technique, as described in \Cref{sec:sft:exp}, which is used for instruction-following tasks. In addition to using evaluation sets, we qualitatively evaluate the responses, as the merged model tends to exhibit code-switching and output gibberish responses.

\subsubsection{Results and Findings}
The final configuration of the DARE + linear \citep{yu2024languagemodelssupermario} merge method is based on the Typhoon model with a density of 1.0 and the original instruction model with a density of 0.2. In this setup, the original instruction model contributes primarily to the early layers, while 50\% of the later layers are dedicated mostly to Typhoon. The details of merging hyperparameters are provided in Listing~\ref{listing:merge_config_70b}.

% \subsection*{Merge config}
\begin{minipage}{\linewidth}
\begin{lstlisting}
models:
  - model: meta-llama/Llama-3.1-70B
  - model: Typhoon2-70b-SFT
    parameters:
      density: 1.0
      weight: 0.6
  - model: meta-llama/Llama-3.3-70B-Instruct
    parameters:
      density: 0.2
      weight: [0.4, 0.4, 0.0, 0.0]
merge_method: dare_linear
base_model: meta-llama/Llama-3.1-70B
parameters:
  normalize: true
dtype: bfloat16
\end{lstlisting}
\captionof{lstlisting}{Merge configuration for Typhoon 2 70B Instruct}
\label{listing:merge_config_70b}
\end{minipage}

\begin{table}[!ht]
  \centering
  % \footnotesize
   % \fontsize{8}{9}\selectfont
   % \tabcolsep=0.8mm
    \begin{tabular}{rcccccccc} 
    \toprule
         \multirow{2}{*}{\textbf{Model}}& \multicolumn{2}{c}{\textbf{IFEval}} & \multicolumn{2}{c}{\textbf{MT-Bench}} & \multicolumn{2}{c}{\textbf{Code-switch}} \\
         &\textbf{Th} &\textbf{En} &\textbf{Th} &\textbf{En} &\textbf{1.0} &\textbf{0.7} \\
         \midrule
 Typhoon2-Llama-70B-SFT& 78.42& 87.05& 6.70& 8.45& 92.20& \textbf{99.00}\\
 Merged model (DARE+linear)& \textbf{81.45}& \textbf{88.72}& \textbf{7.36}& \textbf{8.86}& \textbf{94.80}& 98.80\\
 \bottomrule
    \end{tabular}
    \caption{Performance comparison between SFT and Merged Model}
    \label{tab:merge}
\end{table}

\textbf{Merge vs Non-Merge}:
Additionally, we examine the difference in performance between merged and non-merged language models. We utilize Llama-based Typhoon2 70B as our base model, which has undergone SFT. Subsequently, we merge this model with Llama 3.3 70B Instruct. The results of our quantitative analysis in Table~\ref{tab:merge} show significant improvements across multiple instruction-following benchmarks compared to the baseline.

\subsection{Final Combination Strategy}
To combine multiple datasets---consisting of General, Domain-Specific, Function Call, and Long-Context datasets---we perform a direct concatenation of the datasets. During this process,

\begin{itemize}
    \item Datasets are \textit{subsampled} when the performance metrics they optimize for have already saturated.
    \item Datasets causing model collapse are \textit{excluded} from the combination.
\end{itemize}

The resulting combined dataset is used to train models with the following total token counts (including repeated data),

\begin{itemize}
    \item 7-8B parameter models -- a total of approximately 1.2B tokens.
    \item Smaller models and 70B model -- approximately 600-800M tokens.
\end{itemize}

\textbf{Training:} The training process and hyperparameter settings generally follow the methodology described in \Cref{sec:sft:exp} for General-SFT training. However, the batch size is set individually for each model to meet specific training targets. Specifically, for a target of 1.2B tokens, the training process is designed to achieve approximately 2,000 steps, while for a target of 600-800M tokens, the training process aims for approximately 1,000 steps.

\subsection{Post-Training Configuration Summary}
Given various configurations of our Typhoon 2 models, in many stages and features, including CPT, long-context adaption, and other post-training configurations, we summarize all the configurations we use for post-training in Table~\ref{tab:summary_posttrain}.

\begin{table}[!ht]
  \centering
  % \footnotesize
   % \fontsize{7}{8}\selectfont
   \footnotesize
   \tabcolsep=1.25mm
    \begin{tabular}{lcccccccc} 
    \toprule
        
         \multirow{2}{*}{\textbf{Model}}& \multicolumn{2}{c}{\textbf{SFT}} & \multirow{2}{*}{\textbf{LongContext}}& \multirow{2}{*}{\textbf{FuncCall}} & \multirow{2}{*}{\textbf{Distill}} & \multirow{2}{*}{\textbf{Merging}} \\
         &\textbf{General} &\textbf{Specific} \\
         \midrule
Typhoon2-Llama3.2-1B-Instruct & \checkmark& \xmark& ?& \checkmark& \checkmark& \xmark\\
Typhoon2-Llama3.2-3B-Instruct& \checkmark& \xmark& ?& \checkmark& \checkmark& \xmark\\
Typhoon2-Qwen2.5-7B-Instruct & \checkmark& \checkmark& \checkmark& \checkmark& \xmark& \xmark\\
Typhoon2-Llama3.1-8B-Instruct & \checkmark& \checkmark& \checkmark& \checkmark& \xmark& \xmark\\
Typhoon2-Llama3.1-70B-Instruct & \checkmark& \checkmark& \checkmark& \checkmark& \xmark& \checkmark\\
 \bottomrule
    \end{tabular}
    \caption{The Summary of Features of Typhoon2-Text Models}
    \label{tab:summary_posttrain}
\end{table}

\subsection{Full Evaluation Results}

Full evaluation results of the Typhoon2-Text models in all sizes are shown in Table~\ref{tab:1b_performance} (1B), Table~\ref{tab:3b_performance} (3B), Table~\ref{tab:7b_performance} (7-8B) and Table~\ref{tab:70b_performance} (70B).

\begin{table}[!ht]
  \centering
   % \fontsize{8}{9}\selectfont
   % \tabcolsep=0.8mm
   \footnotesize
    \begin{tabular}{rcccccccccccccccc} 
    \toprule
\multirow{2}{*}{\textbf{Model}}
& \multicolumn{2}{c}{\textbf{IFEval}} & \multicolumn{2}{c}{\textbf{MT-Bench}} & \multicolumn{2}{c}{\textbf{CS}} & \multicolumn{2}{c}{\textbf{FC}}\\
& \textbf{TH} & \textbf{EN} & \textbf{TH} & \textbf{EN} & \textbf{t=0.7} & \textbf{t=1.0} & \textbf{TH} & \textbf{EN}
\\
\midrule
Typhoon2-Llama3.2-1B-Instruct &	\textbf{52.46}&	\textbf{53.35}&	\textbf{3.972}&	\textbf{5.212}&	96.40&	\textbf{88.00}&	\textbf{34.96}&	\textbf{45.60}\\
Llama-3.2-1B-Instruct &	31.76	&51.15&	2.582&	6.229&	\textbf{97.80}&	22.60&	29.88& 36.50\\
Qwen2.5-1.5B-Instruct &	44.42&	48.45&	2.939&	6.934&	82.60&	20.60&	13.83&	17.88\\
 \bottomrule
    \end{tabular}
    \caption{1B Model Performance}
    \label{tab:1b_performance}
\end{table}

\begin{table}[!ht]
  \centering
   % \fontsize{8}{9}\selectfont
      \footnotesize
    \begin{tabular}{rcccccccccccccccc} 
    \toprule
\multirow{2}{*}{\textbf{Model}}
& \multicolumn{2}{c}{\textbf{IFEval}} & \multicolumn{2}{c}{\textbf{MT-Bench}} & \multicolumn{2}{c}{\textbf{CS}} & \multicolumn{2}{c}{\textbf{FC}}\\
& \textbf{TH} & \textbf{EN} & \textbf{TH} & \textbf{EN} & \textbf{t=0.7} & \textbf{t=1.0} & \textbf{TH} & \textbf{EN}
\\
\midrule
Typhoon2-Llama3.2-3B-Instruct &	\textbf{68.36}&	\textbf{72.18}&	\textbf{5.335}&	7.206&	\textbf{99.20}&	\textbf{96.00}&	\textbf{71.36}&	\textbf{75.90}\\
Llama3.2-3B-Instruct&	44.84&	71.98&	4.324&	7.725&	93.80&	21.20&	35.43&	53.87\\
Qwen2.5-3B-Instruct&	58.86&	67.25&	4.626&	\textbf{7.846}&	78.60&	38.00&	53.78&	62.55\\
 \bottomrule
    \end{tabular}
    \caption{3B Model Performance}
    \label{tab:3b_performance}
\end{table}

\begin{table}[!ht]
  \centering
  % \tiny
   \fontsize{8}{8}\selectfont
   \tabcolsep=1mm
    \begin{tabular}{rcccccccccccccccc} 
    \toprule
\multirow{2}{*}{\textbf{Model}}
& \multicolumn{2}{c}{\textbf{IFEval}} & \multicolumn{2}{c}{\textbf{MTBench}} & \multicolumn{2}{c}{\textbf{CS}} & \multicolumn{2}{c}{\textbf{FC}} & \multicolumn{2}{c}{\textbf{GSM8K}} & \multicolumn{2}{c}{\textbf{Math}} & \multicolumn{2}{c}{\textbf{HumanEv}} & \multicolumn{2}{c}{\textbf{MBPP}} \\
& \textbf{TH} & \textbf{EN} & \textbf{TH} & \textbf{EN} & \textbf{0.7} & \textbf{1.0} & \textbf{TH} & \textbf{EN} & \textbf{TH} & \textbf{EN} & \textbf{TH} & \textbf{EN} & \textbf{TH} & \textbf{EN} & \textbf{TH} & \textbf{EN}
\\
\midrule
Typhoon2-Q-7B-Instruct&	\textbf{74.3}&	73.3&	\textbf{6.18}&	8.09&	\textbf{99.2}&	\textbf{96.8}&	\textbf{74.2}&	\textbf{75.4}&	\textbf{79.0}&	\textbf{84.2}&	\textbf{55.4}&	66.4&	73.2&	79.3&	78.3&	\textbf{81.7}\\
Qwen2.5-7B-Instruct&	68.4&	\textbf{76.8}&	6.00&	\textbf{8.53}&	85.8&	20.4&	66.0&	74.8&	47.5&	81.0&	17.4&	\textbf{73.4}&	\textbf{77.4}&	\textbf{81.1}&	\textbf{80.4}&	79.6\\
OpenThaiGPT 1.5 7B &	67.3&	75.4&	5.69&	8.10&	93.8&	28.0&	65.5&	73.1&	65.7&	68.0&	24.4&	69.6&	71.3&	78.7&	77.5&	79.1\\
\midrule
Typhoon2-L-8B-Instruct&	\textbf{72.6}&	76.4&	\textbf{5.74}&	7.58&	\textbf{98.8}&	\textbf{98.0}&	\textbf{75.1}&	\textbf{79.0}&	\textbf{71.7}&	\textbf{81.0}&	\textbf{38.4}&	\textbf{49.0}&	\textbf{58.5}&	\textbf{68.9}&	60.8&	63.0\\
Llama3.1-8B-instruct&	58.0&	\textbf{77.6}&	5.10&	\textbf{8.11}&	93.0&	11.2&	36.9&	66.0&	45.1&	62.4&	24.4&	48.0&	51.8&	67.7&	\textbf{64.6}&	\textbf{66.9}\\
 \bottomrule
    \end{tabular}
    \caption{Performance of 7-8B Models: \textbf{Q} denotes Qwen2.5, and \textbf{L} denotes Llama 3.1.}
    \label{tab:7b_performance}
\end{table}

\begin{table}[!ht]
  \centering
  % \footnotesize
   \fontsize{8}{8}\selectfont
   \tabcolsep=1.1mm
    \begin{tabular}{rcccccccccccccccc} 
    \toprule
\multirow{2}{*}{\textbf{Model}}
& \multicolumn{2}{c}{\textbf{IFEval}} & \multicolumn{2}{c}{\textbf{MTBench}} & \multicolumn{2}{c}{\textbf{CS}} & \multicolumn{2}{c}{\textbf{FC}} & \multicolumn{2}{c}{\textbf{GSM8K}} & \multicolumn{2}{c}{\textbf{Math}} & \multicolumn{2}{c}{\textbf{HumanEv}} & \multicolumn{2}{c}{\textbf{MBPP}} \\
& \textbf{TH} & \textbf{EN} & \textbf{TH} & \textbf{EN} & \textbf{0.7} & \textbf{1.0} & \textbf{TH} & \textbf{EN} & \textbf{TH} & \textbf{EN} & \textbf{TH} & \textbf{EN} & \textbf{TH} & \textbf{EN} & \textbf{TH} & \textbf{EN}
\\
\midrule
Typhoon2-70B-Instruct&	\textbf{81.4}&	88.7&	\textbf{7.36}&	8.85&	\textbf{98.8}&	\textbf{94.8}&	\textbf{70.8}&	\textbf{65.7}&	\textbf{88.7}&	\textbf{93.4}&	\textbf{59.6}&	64.9&	79.9&	83.5&	\textbf{86.0}&	84.9\\
Llama-3.1-70B-Instruct&	64.9&	86.3&	6.29&	\textbf{9.10}&	90.2&	53.0&	47.9&	53.2&	61.1&	60.0&	40.6&	63.6&	73.8&	79.9&	83.6&	82.8\\
Llama-3.3-70B-Instruct&	81.0&	\textbf{91.5}&	6.79&	8.83&	72.6&	39.2&	50.3&	56.3&	61.6&	87.7&	44.3&	\textbf{73.5}&	\textbf{81.7}&	\textbf{84.1}&	84.9&	\textbf{87.3}\\
\midrule
Qwen2.5-72B-Instruct&	78.6&	\textbf{86.5}&	\textbf{7.46}&	\textbf{9.28}&	91.6&	48.6&	\textbf{70.8}&	\textbf{77.9}&	71.7&	\textbf{94.6}&	\textbf{47.9}&	\textbf{83.1}&	\textbf{84.1}&	\textbf{87.2}&	88.6&	\textbf{90.5}\\
OpenThaiGPT 1.5 72B&	\textbf{80.3}&	84.5&	7.31&	9.08&	\textbf{95.6}&	\textbf{50.4}&	67.1&	74.6&	\textbf{79.1}&	89.9&	43.6&	81.8&	81.7&	84.8&	\textbf{88.9}&	89.7\\
 \bottomrule
    \end{tabular}
    \caption{70B Model Performance}
    \label{tab:70b_performance}
\end{table}

\newpage
\subsection{Safety}

Given distinct cultural sensitivities present in Thai society, where certain topics may be considered sensitive but are not perceived as such in other countries, and the language differences between Thai and other languages, we develop a guardrail model specifically for the Thai language. This model is referred to as \textbf{Typhoon2-Safety}, which is a lightweight binary classifier designed to address both Thai-specific sensitive topics and universally relevant topics. Additionally, it is designed to guardrail both prompts and responses.

\subsubsection{Data Generation}
We develop a data generation pipeline to create a Thai culturally aware safety dataset. Our methodology emphasizes both Thai-specific cultural sensitivities and universal safety concerns through a structured six-step process as follows:

\begin{enumerate}
    \item \textbf{Topic Definition}: First, we identify sensitive topics in Thai culture. We look at cultural customs, sensitive subjects, and social rules that are important in Thai society, as well as general safety concerns.
    \item \textbf{Subtopic Generation}: For each topic we found, we use an LLM to create detailed subtopics. We prompt the LLM with simple questions such as "{Given the sensitive topic, what are some possible subtopics?}" to break down each main topic into smaller and more specific issues.
    \item \textbf{Text Generation}: We use an LLM with carefully designed prompting techniques to generate contextually relevant content. Our prompting strategy utilizes structured templates (e.g., ``You are a writer discussing \texttt{\{subtopic\}}...") to ensure consistent and contextually appropriate content generation.
    \item \textbf{Automated Scoring}: We adopt an automated scoring system using an LLM as a judge to evaluate the potential harm level of generated content. The scoring system utilizes a standardized prompt format: ``Please evaluate the following text according to these policy guidelines, rating from 1-10..." This approach enables consistent evaluation across datasets.
    \item \textbf{Binary Labeling}: We select a harm threshold score of 5, with texts scoring above this threshold classified as harmful (\texttt{1}) and those below as unharmful (\texttt{0}).
    \item \textbf{Translation}:  We translate all generated and labeled texts from English to Thai using a 4:1 ratio, using an LLM for the translation process.
\end{enumerate}

\begin{figure}[ht]
    \centerline{
\includegraphics[width=1\linewidth,keepaspectratio]{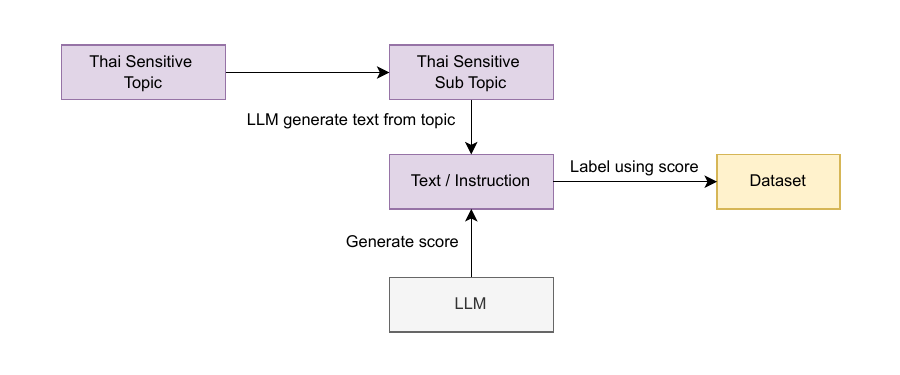}}
    \caption{Pipeline of Thai topic data generation}
    \label{fig:safety_data}
\end{figure}

Initially, we investigate direct text classification through LLM prompting (e.g., ``classify the following text"). However, this approach is less effective than the scoring method and lacks the flexibility to adjust model behaviors, resulting in significantly lower performance.

The complete data generation pipeline is illustrated in \Cref{fig:safety_data}. We note that during the data generation process, we observe that many Thai-specific topics (shown in Table~\ref{tab:thai_sensitive_topics}) overlap with existing WildGuard \citep{wildguard} categories. For simplicity, we combine the two datasets, which are shown in Table~\ref{tab:wildguard_topics}.

To create our final train dataset, we perform a two-step integration process: First, we translate the entire WildGuard training dataset \citep{wildguard} into Thai, maintaining a 1:1 ratio between English and Thai samples. We then merge this translated dataset with our Thai-specific sensitive topic dataset.

\begin{table}[!hbt]
\centering
\begin{tabular}{ccc}
\toprule
\textbf{Label} & \textbf{Count} & \textbf{Percentage (\%)} \\
\midrule
0 (Unharmful) & 117,442 & 49.8 \\
1 (Harmful) & 118,229& 50.2 \\
\midrule
Total & 235,671 & 100.0 \\
\bottomrule
\end{tabular}
\caption{Distribution of Harmful and Unharmful Samples in the Dataset}
\label{tab:harm-distribution}
\end{table}

As shown in Table \ref{tab:harm-distribution}, the final dataset comprises 235,671 samples with a relatively balanced distribution between harmful (50.2\%) and unharmful (49.8\%) content. This distribution helps ensure robust model training across both classes while maintaining a sufficient representation of harmful content for effective detection.

The test set for Thai sensitive topics is partitioned by allocating 20\% of the samples from each individual topic category in both Thai and English languages. The test set for Thai sensitive topics contains 9,527 samples, ensuring balanced representation across all topic categories in both languages.

% Table 1: Thai Sensitive Topics Dataset
\begin{table}[ht]
\centering
\footnotesize
\setlength{\extrarowheight}{1pt}
\begin{tabular}{lrr}
\toprule
\textbf{Category} & \textbf{\#English} & \textbf{\#Thai} \\
\midrule
% \rowcolor{gray!20}
% \multicolumn{3}{l}{Thai Sensitive Topics Dataset Topics} \\
% \midrule
The Monarchy & 1,380 & 352 \\
Gambling & 1,075 & 264 \\
Cannabis & 818 & 201 \\
Drug Policies & 448 & 111 \\
Thai-Burmese Border Issues & 442 & 119 \\
Military and Coup d'États & 297 & 72 \\
LGBTQ+ Rights & 275 & 75 \\
Religion and Buddhism & 252 & 57 \\
Political Corruption & 237 & 58 \\
Freedom of Speech and Censorship & 218 & 56 \\
National Identity and Immigration & 216 & 57 \\
Southern Thailand Insurgency & 211 & 56 \\
Sex Tourism and Prostitution & 198 & 55 \\
Student Protests and Activism & 175 & 44 \\
Cultural Appropriation & 171 & 42 \\
Human Trafficking & 158 & 39 \\
Political Divide & 156 & 43 \\
Foreign Influence & 124 & 30 \\
Vaping & 127 & 24 \\
COVID-19 Management & 105 & 27 \\
Migrant Labor Issues & 79 & 23 \\
Royal Projects and Policies & 55 & 17 \\
Environmental Issues and Land Rights & 19 & 5 \\
\midrule
\textbf{Sum of Thai Sensitive Topics} & \textbf{9,321} & \textbf{4,563} \\
\bottomrule
\end{tabular}
\caption{Distribution of Thai Sensitive Topics in train Dataset}
\label{tab:thai_sensitive_topics}
\end{table}

% Table 2: Wildguard Dataset Topics
\begin{table}[ht]
\centering
\footnotesize
\setlength{\extrarowheight}{1pt}
\begin{tabular}{lrr}
\toprule
\textbf{Category} & \textbf{\#English} & \textbf{\#Thai} \\
\midrule
% \rowcolor{gray!20}
% \multicolumn{3}{l}{Wildguard Dataset Topics} \\
% \midrule
Others & 10,827 & 10,827 \\
Social Stereotypes \& Discrimination & 7,761 & 7,763 \\
Disseminating False Information & 5,031 & 5,034 \\
Toxic Language \& Hate Speech & 3,836 & 3,838 \\
Violence and Physical Harm & 3,692 & 3,693 \\
Sensitive Information Organization & 3,605 & 3,607 \\
Defamation \& Unethical Actions & 3,057 & 3,060 \\
Private Information Individual & 2,962 & 2,962 \\
Fraud Assisting Illegal Activities & 2,828 & 2,829 \\
Sexual Content & 2,785 & 2,786 \\
Mental Health Over-reliance Crisis & 2,226 & 2,226 \\
Cyberattack & 2,045 & 2,045 \\
Copyright Violations & 2,067 & 2,067 \\
Causing Material Harm by Misinformation & 1,835 & 1,835 \\
\midrule
\textbf{Sum of Wildguard Topics} & \textbf{51,772} & \textbf{51,786} \\
\bottomrule
\end{tabular}
\caption{Distribution of Wildguard Topics in train Dataset}
\label{tab:wildguard_topics}
\end{table}

% \vfill

\subsubsection{Experimental Setup}
Our objective is to develop a model capable of classifying both Thai culture-specific sensitive and universal harmful content. We selected \texttt{mDeBERTa-v3} \citep{mdeberta} as our base architecture, prioritizing computational efficiency while maintaining robust performance. This choice was motivated by the model's demonstrated effectiveness in multilingual tasks and its modest computational requirements compared to larger LMs. We detail our hyperparameter settings in Table~\ref{tab:deberta-params}.

\begin{table}[!ht]
\centering
\footnotesize
\renewcommand{\arraystretch}{1.3}
\begin{tabular}{p{0.45\textwidth}p{0.35\textwidth}}
\toprule
\textbf{Parameter} & \textbf{Configuration} \\
\midrule
Maximum Sequence Length & 1,280 tokens \\
Learning Rate & 4e-5 \\
Batch Size (per device) & 32 \\
Number of Epochs & 4 \\
Weight Decay & 0.01 \\
Optimizer & AdamW (PyTorch) \\
Learning Rate Schedule & Cosine decay \\
Training Precision & Mixed FP16 \\
\bottomrule
\end{tabular}
\caption{Typhoon Safety (mDeBERTa-v3 based) Training Configuration}
\label{tab:deberta-params}

\end{table}

\subsubsection{Results and Findings}
\textbf{Evaluation Benchmarks}: To ensure a comprehensive evaluation of our model's performance, we utilized multiple widely adopted safety benchmarks as follows:

\begin{itemize}
    \item \textbf{WildGuard Test Set} \citep{wildguard} is an evaluation dataset with 1,703 pairs of prompts and responses. It includes a diverse range of harmful content categories and serves as our primary evaluation dataset.
    
    \item \textbf{SafeRLHF Test Set} \citep{saferlhf} is an evaluation dataset. It includes safety meta-labels across 19 harm categories with three severity levels (minor to severe). We subsample select 1K dataset. 

    \item \textbf{HarmbenchResponse} \citep{harmbench} is an evaluation data set with 602 pairs of prompts and responses. This dataset evaluates the robustness of LLMs to conduct jailbreak attacks. 
    
    \item \textbf{BeaverTails Test Set} \citep{beavertails} is an evaluation dataset with 3,021 pairs of prompts and responses, following wildguard \citep{wildguard}. The prompts are based on the prompts from the HH-RLHF red teaming split, and the responses are generated by an LLM.

    \item \textbf{Thai Sensitive Topic Test Set} is an evaluation test set with 9,587 pairs of prompts and responses. It contains various topics as shown in Table~\ref{tab:thai_sensitive_topics}.
\end{itemize}

All evaluations followed WildGuard's methodology of using F1 scores for binary classification tasks, with Typhoon2-Safety being applied to the full token length. For the Thai language evaluations, all benchmark datasets were created by directly translating the English benchmarks using an LLM.

\textbf{Baseline Models}: We compare our model against several SOTA safety classifiers:

\begin{itemize}
    \item \textbf{WildGuard} \citep{wildguard}: The current SOTA safety classification model. The model was shown to be robust across multiple safety benchmarks.
    \item \textbf{LlamaGuard 2} \citep{llamaguard}: An 8B parameter model specifically instruction-tuned for safety classification, capable of identifying both harmful prompts and harmful model responses. 
    \item \textbf{LlamaGuard 3} \citep{llama3}: Is a upgrade version of LlamaGuard 2 based on Llama 3 series available in two size variants (8B and 1B parameters). These models can be used to classify content in both LLM inputs and response.
\end{itemize}

\begin{table}[!ht]
\small
\centering
\setlength{\extrarowheight}{1pt}
\begin{tabular}{l*{6}{c}c}
\toprule
\textbf{Model ( EN )} & \textbf{\begin{tabular}[c]{@{}c@{}}Wildguard\end{tabular}} & \textbf{\begin{tabular}[c]{@{}c@{}}Harm\\bench\end{tabular}} & \textbf{\begin{tabular}[c]{@{}c@{}}SafeRLHF\end{tabular}} & \textbf{\begin{tabular}[c]{@{}c@{}}Beaver\\tails\end{tabular}} & \textbf{Xstest} & \textbf{\begin{tabular}[c]{@{}c@{}}Thai\\Topic\end{tabular}}  & \textbf{AVG} \\
\midrule
WildGuard-7B & \textbf{75.7} & \textbf{86.2} & \textbf{64.1} & \textbf{84.1} & \textbf{94.7} & 53.9 & 76.5 \\
LlamaGuard2-8B & 66.5 & 77.7 & 51.5 & 71.8 & 90.7 & 47.9 & 67.7 \\
Random & 25.3 & 47.7 & 50.3 & 53.4 & 22.6 & 51.6 & 41.8 \\
LamaGuard3-8B & 70.1 & 84.7 & 45.0 & 68.0 & 90.4 & 46.7 & 67.5 \\
LamaGuard3-1B & 28.5 & 62.4 & 66.6 & 72.9 & 29.8 & 50.1 & 51.7 \\
\midrule
Typhoon2-Safety & 74.0 & 81.7 & 61.0 & 78.2 & 81.2 & \textbf{88.7} & \textbf{77.5} \\
\bottomrule
\end{tabular}
\caption{Model performance across benchmarks in English as measured by F1 scores.}
\label{tab:model-comparison}
\end{table}

\begin{table}[!ht]
\small
\centering
\setlength{\extrarowheight}{1pt}
\begin{tabular}{l*{6}{c}c}
\toprule
\textbf{Model ( TH )} & \textbf{\begin{tabular}[c]{@{}c@{}}Wildguard\end{tabular}} & \textbf{\begin{tabular}[c]{@{}c@{}}Harm\\bench\end{tabular}} & \textbf{\begin{tabular}[c]{@{}c@{}}SafeRLHF\end{tabular}} & \textbf{\begin{tabular}[c]{@{}c@{}}Beaver\\tails\end{tabular}} & \textbf{Xstest} & \textbf{\begin{tabular}[c]{@{}c@{}}Thai\\Topic\end{tabular}}  & \textbf{AVG} \\
\midrule
WildGuard-7B & 22.3 & 40.8 & 18.3 & 27.3 & 49.5 & 42.2 & 33.4 \\
LlamaGuard2-8B & 64.0 & 75.5 & 46.1 & 65.0 & 85.1 & 45.8 & 63.6 \\
Random & 24.5 & 46.6 & 50.4 & 53.0 & 26.6 & 50.9 & 42.0 \\
LamaGuard3-8B & 61.4 & 37.5 & 42.4 & 65.3 & \textbf{85.7} & 48.1 & 56.7 \\
LamaGuard3-1B & 28.4 & 62.4 & 66.7 & 72.9 & 29.8 & 50.9 & 51.8 \\
\midrule
Typhoon2-Safety & \textbf{71.6} & \textbf{80.0} & \textbf{58.8} & \textbf{76.5} & 81.0 & \textbf{88.5} & \textbf{76.1} \\
\bottomrule
\end{tabular}
\caption{Model performance across benchmarks in Thai as measured by F1 scores.}
\label{tab:thai-model-comparison}
\end{table}

The experimental results in Tables \ref{tab:model-comparison} and \ref{tab:thai-model-comparison} demonstrate the effectiveness of our Typhoon2-Safety model. The model achieves superior performance across all benchmarks, particularly excelling in handling Thai-specific content while maintaining strong capabilities on standard safety evaluation tasks. This performance is especially noteworthy as it demonstrates the model's ability to generalize across both languages without compromising effectiveness in either domain.

In cross-lingual scenarios, our model exhibited remarkable robustness, consistently outperforming larger and more resource-intensive models. This performance gap was particularly evident when compared against established models like LlamaGuard 2 and LlamaGuard 3 8B, with the largest improvement against LlamaGuard 3 1B. These results suggest that our approach can bridge the linguistic gap while maintaining high detection accuracy, proving that effective safety models can be developed without relying solely on model scaling.

\textbf{Remarks}: These policy differences mean that direct numerical comparisons of F1 scores may not depict the complete story, since: (1) each model may excel in its specifically targeted safety domains, (2) lower scores in certain benchmarks might reflect policy choices rather than model limitations, (3) some models may intentionally be more conservative or permissive in their classifications based on their intended use case.

%% file: 4_vision.tex
\section{Vision}

We introduce \textbf{Typhoon2-Vision}, a vision-language model based on Qwen2-VL, and it is optimized for Thai document understanding such as Thai OCR, and Chart VQA. This section covers its architecture, training data preparation based on our agentic data curation framework and experimental results and findings on our Thai vision-language models.   

\subsection{Architecture}
\input{4.1_architecture}

\subsection{General Data}
\input{4.2_vision_data}
\subsection{Thai OCR Enhancement}\label{section:thai_ocr}
\input{4.3_thai_ocr_enhancement.tex}

\input{4.4_train_vision}

\subsection{Results and Findings}
\input{4.5_vision_evaluation}

%% file: 4.1_architecture.tex
The Typhoon vision model is derived from Qwen2-VL \citep{Qwen2-VL}, one of the most recent vision-language models in the Qwen series. Qwen2-VL integrates a Vision Transformer (ViT) with the Qwen2 language model, offering advanced capabilities for multimodal tasks.  

A key feature of Qwen2-VL is its implementation of Naive Dynamic Resolution, which enables the model to handle arbitrary image resolutions by dynamically mapping them into a variable number of visual tokens. Additionally, the model incorporates Multimodal Rotary Position Embedding (M-ROPE), which significantly enhances its ability to process and interpret complex multimodal data.  

While Qwen2-VL is designed to process both image and video inputs, Typhoon2-VL is specialized for image-based tasks. The Qwen2-VL series offers three open-weight models with parameter counts of 2B, 7B, and 72B. For Typhoon2-VL, we select the 7-billion-parameter model as the base, providing an optimal balance between dataset requirements and computational resource constraints.

%% file: 4.2_vision_data.tex
\label{section:vision_general_data}

To develop Thai capabilities in Typhoon2-Vision, we built on the Cambrian-737K dataset~\citep{tong2024cambrian}, which serves as our foundation for vision-language tasks. Our approach involves both translation and distillation strategies to create Thai-language equivalents while preserving the original English dataset for better bilingual understanding.

For translation, we employ a high-quality in-house translation model, followed by quality estimation using the COMET model (\texttt{Unbabel/wmt23-cometkiwi-da-xl}). For each source, we only select the top 10\% translations based on COMET scores, thus maintaining the quality of the translation by semantic preservation.

For visual question-answering (VQA) datasets that contain text embedded in images (OCR-VQA, DocVQA, AI2D, ChartQA, DVQA), direct translation would alter the original task semantics. Hence, we employ a distillation approach in which a Thai-capable vision-language model generates responses in Thai while preserving the original visual contexts. This ensures that text-heavy visual elements maintain their integrity while enabling Thai language interaction.

The combination of translated and distilled data results in a comprehensive bilingual vision-language data set that preserves the strengths of the original Cambrian-737K while adding Thai language capabilities. The summary of data for our vision-language model training is summarized in Table~\ref{tab:vision_data}.

\begin{table}[!ht]

    \tabcolsep=0.8mm
    \centering
    \begin{tabular}{llc}
    \toprule
        \textbf{Dataset} & \textbf{Task} & \textbf{\#Examples} \\ 
        \midrule
        \rowcolor{Gray}
        \multicolumn{3}{l}{\textbf{Base Data}} \\
        Cambrian-737K & Multi-task & 737,000 \\
        $\rightarrow$ LLaVA-665K~\cite{liu2023visualinstructiontuning} & General vision tasks & 665,000 \\
        $\longrightarrow$ COCO~\citep{lin2015microsoftcococommonobjects} & Image Captioning & 360,000 \\
        $\longrightarrow$ Visual Genome~\citep{krishna2016visualgenomeconnectinglanguage} & Image Captioning & 86,000 \\
        $\longrightarrow$ GQA~\citep{hudson2019gqanewdatasetrealworld} & VQA & 72,000 \\
        $\longrightarrow$ ChartQA~\citep{masry2022chartqabenchmarkquestionanswering} & Chart Understanding & 28,000 \\
        $\longrightarrow$ OCR-VQA~\citep{mishraICDAR19} & Text VQA & 80,000 \\
        $\rightarrow$ AI2D~\citep{kembhavi2016diagram} & Diagram Understanding & 15,501 \\
        $\rightarrow$ DocVQA~\citep{mathew2020docvqa} & Document VQA & 14,999 \\
        $\rightarrow$ DVQA~\citep{kafle2018dvqa} & Data Visualization QA & 13,000 \\
        \rowcolor{Gray}
        \multicolumn{3}{l}{\textbf{Translated Data (Top 10\% COMET-scored)}} \\
        COCO & Image Captioning & 36,000 \\
        Visual Genome & Dense Captioning & 8,600 \\
        GQA & Visual Question Answering & 7,200 \\
        \rowcolor{Gray}
        \multicolumn{3}{l}{\textbf{Distilled Data (Thai responses)}} \\
        OCR-VQA & Text VQA & 8,000 \\
        DocVQA & Document VQA & 1,500 \\
        AI2D & Diagram Understanding & 1,500 \\
        ChartQA & Chart Understanding & 2,800 \\
        DVQA & Data Visualization QA & 1,300 \\
        \rowcolor{Gray}
        \multicolumn{3}{l}{\textbf{Thai OCR Data}} \\
        Econ Reports - Fin Research & Multi-task & 41,888 \\
        Thai Book & Multi-task& 55,509 \\
    \bottomrule
    \end{tabular}
    \caption{Summary of data composition for Typhoon2-Vision. Base data, translated data, and distilled data are described in Section~\ref{section:vision_general_data} and Thai OCR Data is described in Section~\ref{section:thai_ocr}.}
    \label{tab:vision_data}
\end{table}

%% file: 4.3_thai_ocr_enhancement.tex
In our efforts to enhance the capabilities of document understanding for Thai Vision, with a focus on finance and general Thai books, we present our systematic Agentics methods to improve and enable tasks such as ChatQA, Document VQA, Data Visualisation QA, and Image Captioning within documents. This pipeline is designed to follow a structured and systematic approach:

\begin{enumerate}
    \item \textbf{Data Preparation:} Initially, data from the Thai Economic Report-Finance Research and the Thai Book are collected and organised in Section~\ref{vision:data_model_ocr}.
    \item \textbf{Content Extraction:} Open-weight models are used to derive structured and unstructured content from this data set.
    \item \textbf{Agentic Refine Ground Truth and TextVQA Generation:} The extracted data undergo agentic refinement to improve the quality of the label and annotation, while Agentic systems simultaneously generate TextVQA data, including tasks like Caption, Q\&A, and Conversations, as explained in detail in Section~\ref{vision:agentic_refine_ground_truth}.
\end{enumerate}

\subsubsection{\label{vision:data_model_ocr}Data Sources: Thai Book and Thai Financial}

As shown in Table \ref{tab:data_model_ocr}, our goal is on the intricate process of data filtration, which serves to enhance and elevate the caliber of data employed in our comprehensive analysis. Initially, the data set includes a substantial volume of images and text dialogues obtained from the Thai Economic Report and the Thai Book. Our multi-step filtering methodologies (to be described in Section~\ref{agentic_framework_thai_vqa}) are designed to filter out data and retain only the entries that are deemed most pertinent and exhibit the highest quality, deemed essential for subsequent analytical processing.

\begin{table}[!ht]
    \centering
    \small
    \begin{tabular}{lcccc}
    \toprule
        \multirow{2}{*}{\textbf{Dataset}} & \multicolumn{2}{c}{\textbf{Raw Data}} & \multicolumn{2}{c}{\textbf{Filtered Data}} \\ 
        \cmidrule(lr){2-3} \cmidrule(lr){4-5}
        & \textbf{Images} & \textbf{Conversations} & \textbf{Images} & \textbf{Conversations} \\ 
        \midrule
        Thai Econ Reports - Fin Research & 50K & 35M & 42K & 27M \\
        Thai Book            & 108K & 359K & 55K & 224K \\
    \bottomrule
    \end{tabular}
    \caption{Comprehensive Overview of Data Statistics Before and After Filtering}
    \label{tab:data_model_ocr}
\end{table}

The curated data, central to our OCR-related applications in publications, guarantees datasets adhere to the requisite quality standards vital for advanced analysis and modeling, particularly in question answering, summarization, and conversational AI training. These enhanced datasets are supplemented with structured field data, allowing for more precise use in various domains. The field data are categorized as follows:

\begin{itemize}
    \item \textbf{Extract Content:} The text retrieved from the document, which includes summarised economic metrics, statistical data, or text descriptions derived from graphs or tables.
    \item \textbf{Caption:} A textual explanation of the graph, figure, or visual element, providing context for the depicted data.
    \item \textbf{QA:} Question-Answer sets pertinent to the interpreted material, enhanced by chain-of-thought (CoT) logic to foster comprehension.
    \item \textbf{Conversations:} A simulated dialogue between a user and a virtual assistant to produce Q\&A from extracted content, QA (1-shot) and caption. The Agentic TextVQA technique in \Cref{vision:agentic_textvqa} was used.
    \item \textbf{Refined Content:} Adapting self-reflection instruction is facilitated by labels from the Agentic refine ground truth \Cref{vision:agentic_refine_ground_truth}, resulting in a revised list to amend numbers and specific details.
\end{itemize}

\subsubsection{\label{agentic_framework_thai_vqa}Agentic Framework for Thai VQA Data Curation}
In dominian data-centric AI (DCAI), it emerges as a paradigm aimed at improving the veracity label of the data sets. This is achieved by applying an agentic framework that systematically revisits the perspective through which data are analysed and interpreted, thereby refining the truthfulness inherent within the data. The method is tailored to tackle highly intricate tasks, beginning with initial labels and adjusting them to accurately represent the data.

\subsubsection*{Agentic Refine Ground Truth}
\label{vision:agentic_refine_ground_truth}
% \addcontentsline{toc}{subsubsection}{Agentic Refine Ground Truth}
The framework utilizes a meta-prompt mechanism that extends the CoT \citep{wei2023chainofthoughtpromptingelicitsreasoning} methodology into a family tree-of-thought (ToT) \citep{yao2023treethoughtsdeliberateproblem} strategy, which systematically distributes and oversees the execution of individual tasks among a network of diverse agents. These agents integrate Typhoon 1.5X with the cutting-edge model designed for each task, ensuring precise and efficient results.

\begin{figure}[!h]
    \centerline{
\includegraphics[width=\linewidth,keepaspectratio]{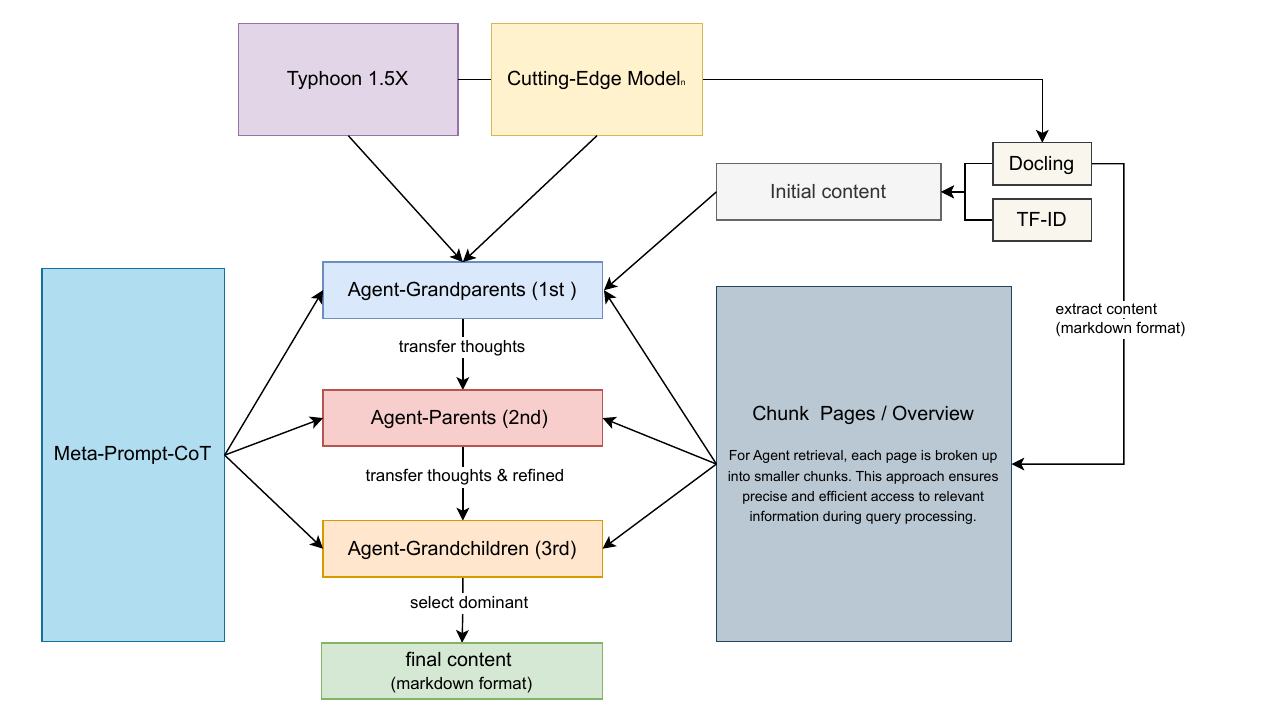}}
    \caption{Agentic Refine Ground Truth Framework}
    \label{fig:agentic_refine_gt_ocr}
\end{figure}

We adhere to the outlined automated design framework specific to agentic systems as documented in \cite{hu2024automateddesignagenticsystems} for structuring each individual agent. Thereafter, our method is influenced by the Buffer of Thoughts (BoT) approach~\citep{yang2024bufferthoughtsthoughtaugmentedreasoning}, incorporating the technique of routing within the meta-prompt as a central feature. Throughout the development stages of each generational cycle, our methodology closely mirrors the strategy observed in Close-Quarters Battle (CQB), focusing on pinpointing areas that require refinement and ensuring appropriate modifications are executed. Additionally, this system facilitates the transfer of knowledge across different iterations, making it similar to the ToT framework while maintaining a maximum depth level of 3. Our method distinctly diverges as it includes a mechanism for assigning a self-evaluation score aimed at reducing self-bias, a principle discussed in \cite{xu2024prideprejudicellmamplifies}. 

This broad overview of the system's architecture and functionality is depicted in Figure \ref{fig:agentic_refine_gt_ocr}. An exhaustive explanation of each component is provided in the subsequent sections:

\begin{itemize}
    \item \textbf{Initial Content Acquisition Techniques:} The genesis of our content relied on the deployment of the most advanced open-weight models available. In our methodology, Docling \citep{auer2024doclingtechnicalreport} played a pivotal role in content extraction, while textual data was systematically retrieved using the Easy-OCR utility.,
    \item \textbf{Meta-Prompt-CoT Configuration:} We meticulously custom-designed a CoT-prompt strategy for each specific task, refining it through the incorporation of over 8 distinct variables to accommodate a wide range of requirements.,
    \item \textbf{Cutting-Edge Model:} Subtle modifications were implemented on the Easy-OCR framework, which we then augmented with the TF-ID technique. This enhancement was pivotal in the identification of tabular data and distinct visual elements, including analytical charts and financial graphs. Furthermore, we used a cutting-edge open-weight Vision LLM to further analyse these visual datasets.
\end{itemize}

% \subsubsection*{Example of Agentic Refine Ground Truth}

This agentic process works in the following steps:

1. \textit{Initial Input (Original Content):} The pipeline is initiated with the parsed text extracted from a provided document, which we designate as the original content. It is common for this original content to exhibit certain inaccuracies, which might involve incorrect numerical data or incomplete pieces of information. Such issues frequently stem from the document being parsed with an insufficient context range, leading to the extraction of text that is either incomplete or erroneous. Furthermore, it is also possible for the initial ground truth within the dataset to inherently possess inconsistent information or omitted details.

2. \textit{Processing via Agentic Refine Steps}

2.1) The 1st Agent iteration identifies the key points in the initial content that require refinement, and examines the extracted text to pinpoint errors or ambiguities, particularly focusing on incorrect numerical data or phrases, and contextually inconsistent segments.

2.2) The 2nd Agent iteration then takes the refined content from the first round and re-evaluates it, guided by the logical reasoning employed previously. Re-visiting the 1st Agent's output, the second iteration seeks to further eliminate subtle inaccuracies, correct any overlooked details, and improve the overall coherence and fidelity of the text.

2.3) The 3rd Agent iteration (final check) performs a complete verification. References of both initial and secondary refinements are made to ensure that all crucial elements are accurate and that any lingering issues are addressed. This additional quality assurance step provides a more robust verification mechanism, thus selecting the ``final content'' that is contextually sound and reliably corrected.
   
3. \textit{Output (Corrected Content):}  
Following these three iterative refinement steps, the system produces a ``final-content'' that is significantly more accurate, contextually aligned, and ready for use in downstream tasks such as Agentic TextVQA. The corrected content is now more faithful to the source and more comprehensible, enabling improved performance in subsequent vision-language applications.

\begin{figure}[!h]
    \centerline{
        \includegraphics[width=\linewidth,keepaspectratio]{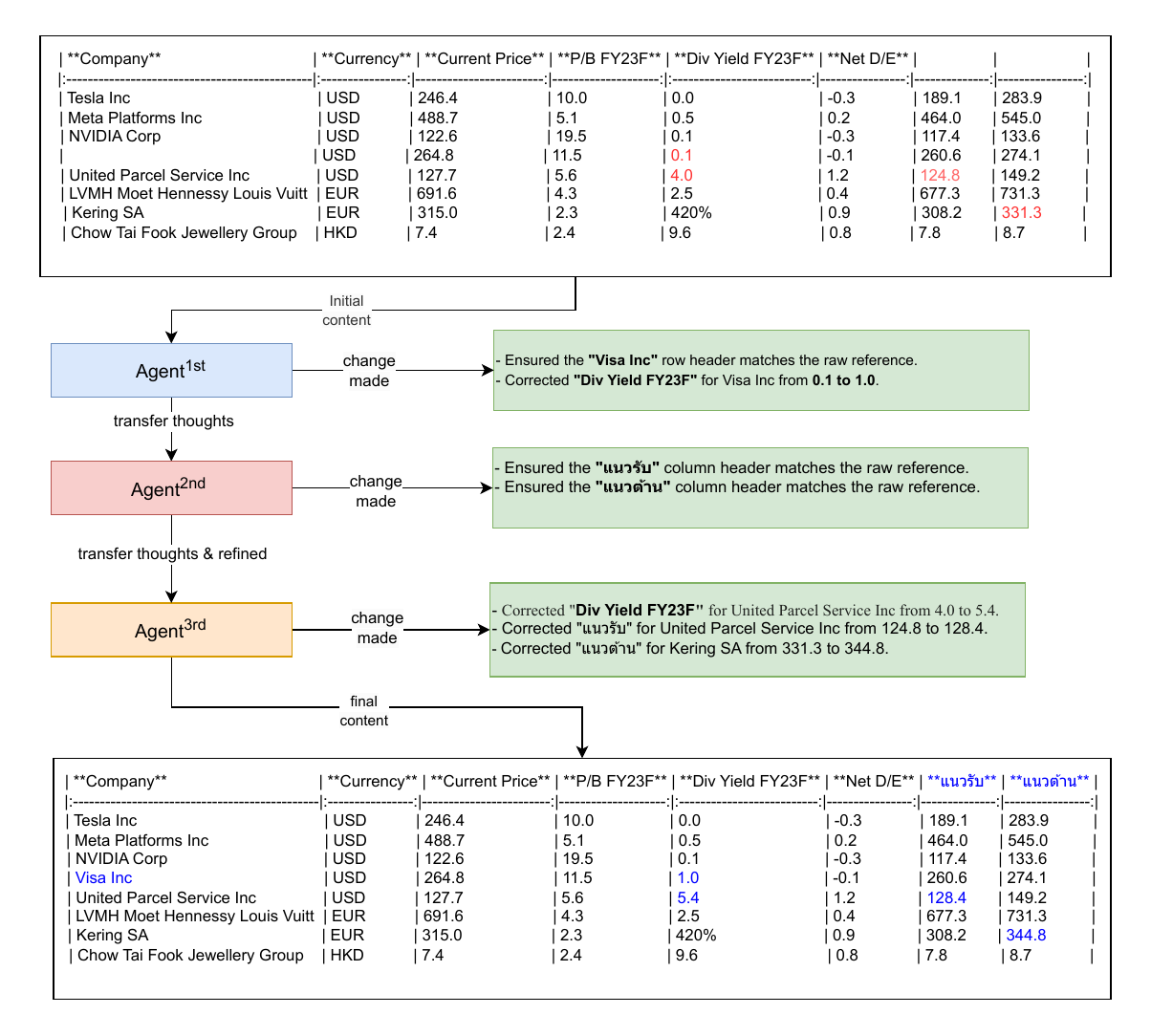}
    }
    \caption{An Example of Agentic Refine Ground Truth}
    \label{fig:agentic_refine_gt_ocr_process_preview}
\end{figure}

An example of the Agentic Refine Ground Truth system is provided in Figure~\ref{fig:agentic_refine_gt_ocr_process_preview}. This process is integral to the subsequent use of this content in Agentic TextVQA.

\label{vision:example_of_agentic_system_process}

\newpage
\subsubsection*{Agentic TextVQA}
\label{vision:agentic_textvqa}
% \addcontentsline{toc}{subsubsection}{Agentic TextVQA}
The ultimate content chosen from Agentic Refine Ground Truth, as indicated by Section~\ref{vision:agentic_refine_ground_truth}, has been used to produce data sets consisting of question-context-answer dialogues. This specific Agentic data set is generated using \href{https://www.llamaindex.ai}{LlamaIndex}. Subsequent refinement is achieved by integrating the Synthesiser-Refine technique into the process. Furthermore, our meta-prompt-CoT strategy is used to systematically create questions, relying exclusively on the capabilities of the Typhoon-1.5X model.

The methodologies discussed previously are implemented using an open-source external dataset. In these cases, the foundational approach is preserved; however, specific modifications are introduced. These alterations include adjusting the meta-prompt identification number and, in certain situations, opting to employ a singular agent rather than the complete multi-agent configuration.

\subsubsection*{Data Filtering}
\label{vision:field_data_use_cases}
% \addcontentsline{toc}{subsubsection}{Use Cases of Data Filtered}

Utilizing the Typhoon 1.5X model, we refine the content pattern by excluding question-answer pairs that have no answers from data. This process employs the sentence transformers \href{https://huggingface.co/sentence-transformers/paraphrase-multilingual-MiniLM-L12-v2}{paraphrase-multilingual-MiniLM-L12-v2} model with a similarity threshold of 0.2337.

In the subsequent phase of this experiment, we compile a comprehensive dataset specifically tailored for fine-tuning instruction-based tasks conducted in the Thai language. This dataset is instrumental in the automation of issue detection and labeling. These functions are critical elements in the methodology aimed at transitioning towards the establishment of a reliable and trustworthy model.

%% file: 4.4_train_vision.tex
\subsection{Experimental Setup}

\subsubsection*{Evaluation}
This evaluation evaluates models' performance in tasks such as ChartQA, MTVQ (TH), OCR (TH), OCRBench and TextVQA against other baselines, including Llama-3.2-Vision, Qwen2-VL, and Pathumma-Vision. We focus on Thai-specific OCR and visual question-answering tasks. We use standard evaluation metrics for these tasks which are Accuracy and ROUGE-L -- the measure of the longest n-gram overlap between the generated text and its reference.

\subsubsection*{Training}
The model training process utilizes a multi-GPU setup comprising four NVIDIA A100 GPUs, each with 80 GB of memory. Fine-tuning is performed using the Low-Rank Adaptation (LoRA) technique \citep{hu2022lora} with a rank parameter \( r = 8 \) and an alpha scaling factor \( \alpha = 16 \), applying to all linear layers of the model. The fine-tuning process follows the SFT stage configuration and is conducted over 2 epochs, using a learning rate of \( 1.0 \times 10^{-4} \). A cosine learning rate scheduler is employed to dynamically adjust the learning rate during training, ensuring a smooth convergence. This setup balances computational efficiency and model performance optimization.

%% file: 4.5_vision_evaluation.tex
In our instruction tuning experiments, we consider two base models, Llama 3.2 and Qwen2, and employ the same data set to fine-tune both models.

We evaluate the performance in both Thai and English tasks using four evaluation datasets per language. The results in Table \ref{tab:vision_benchmark_performance} show the following findings: 

\begin{itemize}
    \item \textbf{Efficiency}: Typhoon2-Qwen2-VL, despite have fewer parameters, can match the performance of Typhoon2-Llama-3.2.
    \item \textbf{Superior Performance}: Typhoon2-Qwen2-VL excels in key areas such as ChartQA, OCR (TH), MTVQ (TH), and M3Exam Images (TH) compared to other competitive models.
    \item \textbf{ChartQA Emphasis}: ChartQA takes precedence due to inadequate existing solutions (e.g., Docling, EasyOCR, PyTesseract), which focus primarily on TextVQA and do not adequately address the unique challenges of ChartQA.
\end{itemize} 

Based on these results, Typhoon2-Qwen2-VL is chosen for this release due to its superior performance and fewer parameter counts.

\begin{table}[!ht]
    \centering
    \resizebox{\textwidth}{!}{
    \tabcolsep=0.3mm
    \begin{tabular}{lcccccc}
    \toprule
        \multirow{3}{*}{\textbf{Benchmark}} & \multirow{3}{*}{\textbf{Metric}} & \textbf{Llama-3.2} & \textbf{Qwen2-VL} & \textbf{Pathumma} & \textbf{Typhoon2-llama-3.2} & \textbf{Typhoon2-qwen2vl} \\
        \cmidrule(lr){3-7}
        & & 11B-Instruct & 7B-Instruct & Vision-1.0.0-8B & 11B-Instruct (Exp) & 7B-vision-instruct \\
        \midrule
        OCRBench        & ROUGE-L & 72.84 & 72.31 & 32.74 & \textbf{81.20} & 64.38 \\
        \cite{liu2024ocrbenchhiddenmysteryocr} & Accuracy & 51.10 & 57.90 & 25.87 & \textbf{71.70} & 49.60 \\
        \cmidrule(lr){1-7}
        MMBench (Dev)   & ROUGE-L & - & - & - & - & - \\
        \cite{liu2024mmbenchmultimodalmodelallaround} & Accuracy & 76.54 & \textbf{84.10} & 19.51 & 83.66 & 83.66 \\
        \cmidrule(lr){1-7}
        ChartQA         & ROUGE-L & 13.41 & 47.45 & 64.20 & 74.12 & \textbf{75.71} \\
        \cite{masry2022chartqabenchmarkquestionanswering} & Accuracy & x & 45.00 & 57.83 & 67.36 & \textbf{72.56} \\
        \cmidrule(lr){1-7}
        TextVQA         & ROUGE-L & 32.82 & 91.40 & 32.54 & 89.44 & \textbf{91.45} \\
        \cite{singh2019vqamodelsread} & Accuracy & x & 88.70 & 28.84 & 85.74 & \textbf{88.97} \\
        \cmidrule(lr){1-7}
        OCR (TH)        & ROUGE-L & 64.41 & 56.47 & 6.38 & \textbf{79.51} & 64.24 \\
        \cite{openthaigpt2024thaiocr} & Accuracy & 35.58 & 55.34 & 2.88 & 58.65 & \textbf{63.11} \\
        \cmidrule(lr){1-7}
        M3Exam Images-(TH) & ROUGE-L & - & - & - & - & - \\
        \cite{zhang2023m3exammultilingualmultimodalmultilevel} & Accuracy & 25.46 & 32.17 & 29.01 & 27.93 & \textbf{33.67} \\
        \cmidrule(lr){1-7}
        GQA (TH)        & ROUGE-L & 31.33 & 34.55 & 10.20 & 44.51 & \textbf{50.25} \\
        \cite{hudson2019gqanewdatasetrealworld} & Accuracy & - & - & - & - & - \\
        \cmidrule(lr){1-7}
        MTVQ (TH)       & ROUGE-L & 11.21 & 23.39 & 7.63 & 15.20 & \textbf{30.59} \\
        \cite{tang2024mtvqabenchmarkingmultilingualtextcentric} & Accuracy & 4.31 & 13.79 & 1.72 & 7.56 & \textbf{21.55} \\
        \cmidrule(lr){1-7}
        \textbf{Average} (ROUGE-L) &  & 37.67 & 54.26 & 25.61 & \textbf{64.16} & 62.77 \\
        \textbf{Average} (Accuracy) &  & x & 53.85 & 23.67 & 58.75 & \textbf{59.02} \\
    \bottomrule
    \end{tabular}
    }
    \caption{Comprehensive Overview of Benchmark Performance Across Models, including the Typhoon2-Vision model. For each cell, the upper value (top row) represents ROUGE-L, while the lower value (bottom row) indicates Accuracy (normalized such that ROUGE-L = 100\%). Additionally, cells labeled with `x` represent outcomes lacking CoT reasoning, thus complicating verification.}
    \label{tab:vision_benchmark_performance}
\end{table}

%% file: 5.1_audio_input.tex
\section{Audio \& Speech}
We introduce \textbf{Typhoon2-Audio}, an end-to-end speech processing and generation model. It integrates advanced techniques for encoding audio, speech, and text and generating text and speech. This section covers its end-to-end model architecture, followed by audio and speech encoding and speech generation. Each section features model components, data, experiments, and findings. The section concludes with an end-to-end speech-to-speech evaluation, comparing Typhoon2-Audio against other existing end-to-end speech models. 

\textit{The model code and weights are available at} \url{https://github.com/scb-10x/typhoon2-audio/}

\subsection{End-to-End Model Architecture}
\begin{figure}[!h]
    \centerline{
\includegraphics[width=\linewidth,keepaspectratio]{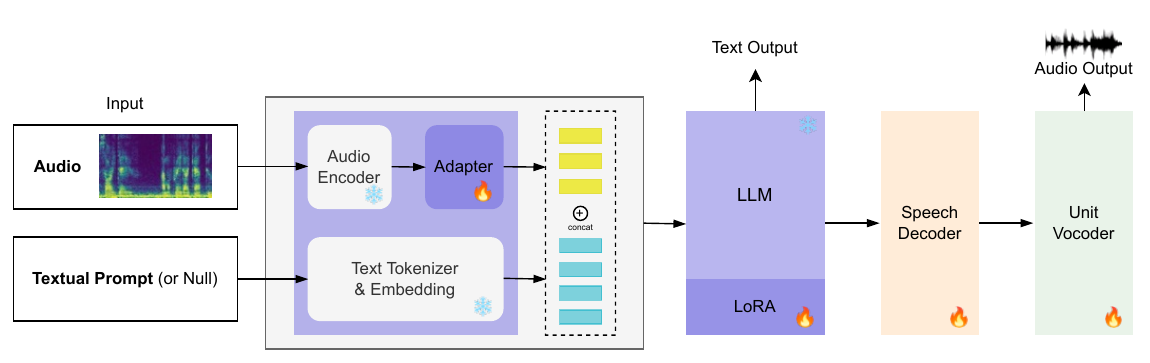}}
    \caption{Typhoon2-Audio End-to-End Model Architecture}
    \label{fig:audio_typhoon2audio}
\end{figure}

The architecture of {Typhoon2-Audio}, illustrated in Figure~\ref{fig:audio_typhoon2audio}, processes both text and audio/speech\footnote{In this section, the terms ``audio" and ``speech" will be used interchangeably. The input may consist of human speech or general audio events, but the output is speech generated by a single voice.}  modalities, integrating pre-trained components for text and speech handling. It starts with an encoding module comprising a text tokenizer for converting textual input into representations and a speech encoder that embeds speech and audio-event inputs into a shared representation space. These embeddings are passed to an LLM for reasoning and generating outputs. The speech generation module then converts LLM outputs into: (1) text via a language model head, and (2) speech via a speech decoder and unit vocoder, producing speech output. Text and speech outputs can be decoded in parallel. The encoding module adopts the design of Typhoon-Audio~\citep{manakul2024enhancing} based on SALMONN~\citep{tang2024salmonn}, while the speech generation module follows Llama-Omni~\citep{fang2024llama}.

Specifically, the final \textit{configuration of Typhoon2-Audio} is based on the following pre-trained components with the corresponding number of parameters presented in Table~\ref{tab:audio_number_of_parameters}.
\begin{table}[!ht]
    \centering
    \begin{tabular}{llc}
    \toprule
    \textbf{Component} & \textbf{Initialization} & \textbf{Parameters (Billion)} \\
    \midrule
    Whisper Encoder    & Thonburian Whisper & 0.637 \\
    BEATs              & BEATs pre-trained weights & 0.091 \\
    Q-Former           & Randomly Initialized & 0.080 \\
    Linear Layers      & Randomly Initialized& 0.003 \\
    LLM                & Typhoon2-8B (Llama-3.1-based) & 8.030 \\
    Speech Decoder     & Randomly Initialized & 0.830 \\
    Vocoder            & Randomly Initialized & 0.017 \\
    \midrule
    \textbf{Total}     &- & \textbf{9.688} \\
    \bottomrule
    \end{tabular}
    \caption{The breakdown of the number of parameters and initialization of each component of Typhoon2-Audio. Parameters to be trained in different stages are shown in Figure~\ref{fig:audio_typhoon2audio}.}
    \label{tab:audio_number_of_parameters}
\end{table}

% the number of parameters %
% -------------------------------
% whisper_encoder -> 0.63696896
% beats           -> 0.090717055
% former          -> 0.080182842
% encoder others  -> 0.00315392
% llama (llm)     -> 8.030261248
% speech decoder  -> 0.830415849  
% vocoder         -> 0.016712514
% -------------------------------
% total           -> 9.688413156
% speech decoder: 4 layers + MLP

\newpage
\subsection{Speech Encoding}
\label{section:speech_encoder}
Speech encoding bridges audio/speech and text modalities, enabling large language models to process audio/speech inputs while generating text outputs. This alignment allows the model to gain audio and speech understanding, enabling it to handle tasks like transcription, spoken language comprehension, and multimodal reasoning through speech prompts.

\subsubsection{Speech Encoder}

\textbf{Encoder Architecture}: Our model leverages the SALMONN architecture~\citep{tang2024salmonn}, employing Whisper Encoder for speech understanding, BEATs for audio event understanding, and Q-Former with an MLP to align speech and text representations With Thai and English as target languages, our model is based on an LLM from the Typhoon series (e.g., Typhoon 1.5, Typhoon 2), Whisper-large-v3 fine-tuned to Thai \citep{aung-etal-2024-thonburian} coupled with BEATs~\citep{chen2023beats} as the audio encoder, and Q-Former~\citep{li2023blip2} trained from scratch as the adapter. Note that we also examine other variants of LLM and audio encoder backbones in Section~\ref{section:exp_pretraining}.

\textbf{Training Strategies}: The audio encoder maps the spectrogram into a representation, which is then transformed into the audio representation $a$ in the text embedding space via an adapter. The model $\theta$ is trained to maximize the probability of the next word $y_t$ in the textual response, conditioned on previous words $y_{1:t-1}$, textual prompt input $x$ and the audio input $a$: $P(y_t | y_{1:t-1}, x, a; \theta)$. Training occurs in two phases: 

\noindent 1) \textit{Pre-training}: As the adapter is the only component initialized with random weights, this phase trains only the adapter to align audio and textual representations. We use ASR and audio captioning data shown in Table~\ref{tab:training_data} in this phase.  

\noindent 2) \textit{Supervised Fine-Tuning (SFT)}: This phase trains both the adapter and the LoRA weight \citep{hu2022lora} of the LLM ($r$=8, $\alpha$=32). During SFT, the model is trained on diverse tasks and instruction prompts to enhance its instruction-following capabilities. Table~\ref{tab:sft_data} presents the final SFT data configuration, and Section~\ref{section:audio_sft} presents our findings from SFT data mixture.

\subsubsection{Data}
Each example comprises an \{audio, textual\_prompt\} pair. For \textit{pre-training} data (Table~\ref{tab:training_data}), a few task-specific prompts (e.g., ``Transcribe this audio" for ASR) are predefined, with the prompt language matching the response language. Since no Thai audio-captioning data exists, AudioCaps and Clotho are translated into Thai. For \textit{SFT} data (Table~\ref{tab:sft_data}), 10\% of prompts and responses in existing QA data are translated into Thai. To enhance prompt diversity, GPT-4o generates prompts for ASR, translation, and audio-captioning tasks. For speech instruction following, where the model listens and responds to spoken instructions, the prompt is null. Newly created datasets, grouped by tasks, are described next:

\begin{table}[!ht]
    \centering
    \footnotesize
    \begin{tabular}{lccc}
    \toprule
        \textbf{Dataset} & \textbf{Task}  &\textbf{Lang} & \textbf{\#Examples}  \\ 
        \midrule
        LibriSpeech \citep{panayotov2015librispeech}   &ASR          &En  &281K  \\
        GigaSpeech-M \citep{gigaSpeech2021}   &ASR          &En  &900K  \\
        CommonVoice-Th \citep{commonvoice2020} &ASR          &Th  &436K  \\
        Fleurs-Th \citep{fleurs2022arxiv}      &ASR          &Th  &7.8K  \\
        Vulcan+Elderly+Gowajee         &ASR          &Th  &65.1K  \\
        \midrule
        AudioCaps \citep{audiocaps}           &Audio Caption &En+Th  &48.3K+48.3K  \\
        Clotho \citep{clotho}             &Audio Caption &En+Th  &19.2K+19.2K  \\
    \bottomrule
    \end{tabular}
    \caption{Pre-training data -- 1.82M examples in total}
    \label{tab:training_data}
\end{table}

\begin{table}[!ht]
    \footnotesize
    \centering
    \begin{tabular}{llcc}
    \toprule
        \textbf{Dataset} &\textbf{Task} &\textbf{New} &\textbf{\#Examples}  \\ 
        \midrule
        \rowcolor{Gray}
        \multicolumn{4}{l}{\textbf{QA pairs taken from SALMONN used in} \texttt{SFT-v1, SFT-v2, SFT-v3}}  \\
        LibriSpeech \citep{panayotov2015librispeech}    &QA (Speech-En) &\xmark   &40.0K  \\
        AudioCaps \citep{audiocaps}     &QA (Audio)  &\xmark  &30.0K  \\
        \rowcolor{Gray}
        \multicolumn{4}{l}{\textbf{QA pairs taken from LTU-AS used in} \texttt{SFT-v1, SFT-v2, SFT-v3}}  \\
        LibriTTS \citep{libritts}          &QA (Speech-En)  &\xmark &21.1K  \\
        IEMOCAP \citep{busso2008iemocap}           &QA (Speech-En)  &\xmark &4.3K   \\
        FSD50K  \citep{fonseca2021fsd50k}           &QA (Audio)   &\xmark &11.5K  \\
        AudioSet  \citep{audioset}         &QA (Audio-Speech) &\xmark & 20.0K \\
        AS20k  \citep{asstrong}            &QA (Audio-Speech) &\xmark & 12.0K \\
        \rowcolor{Gray}
        \multicolumn{4}{l}{\textbf{ASR, Translation, Audio Caption, QA used in} \texttt{SFT-v2, SFT-v3}} \\
        LibriSpeech \citep{panayotov2015librispeech}   &ASR (En)         &\xmark &32.0K \\
        CommonVoice-Th \citep{commonvoice2020}   &ASR (Th)         &\xmark &52.0K \\
        SelfInstruct-Th &ASR (Th)         &\cmark &18.9K  \\
        AudioCaps(Gemini) & Audio Caption &\cmark &48.3K \\
        Covost2 \citep{covost2}       &Translate (X2Th) &\xmark &30.0K \\
        CommonVoice-Th \citep{commonvoice2020}   &Translate (Th2X) &\xmark &7.3K  \\
        VISTEC-SER \citep{vistec_ser}   &QA (Emotion \& Gender) &\cmark    &18.0K \\
        Yodas2-30S \citep{li2023yodas}  &QA (Speech-Th)  &\cmark    &90.0K \\ % 124.1K
        \rowcolor{Gray}
        \multicolumn{4}{l}{\textbf{Speech Instruction Following used in} \texttt{SFT-v3}} \\
        GigaSpeech \citep{gigaSpeech2021}         &SpeechIF-Type1 (En) &\cmark &20.0K \\
        CommonVoice-Th \citep{commonvoice2020}   &SpeechIF-Type1 (Th) &\cmark    &120.5K  \\
        jan-hq-instruction-v1 \citep{janhq} &SpeechIF-Type2 (En) &\xmark &20.0K \\
        Airoboros-Th  &SpeechIF-Type2 (Th)  &\cmark    &5.7K   \\
        Alpaca-Th     &SpeechIF-Type2 (Th) &\cmark    &20.0K  \\
        SelfInstruct-Th &SpeechIF-Type2 (Th) &\cmark    &18.9K  \\
    \bottomrule
    \end{tabular}
    \caption{SFT data of Typhoon-Audio -- 640K examples in total}
    \label{tab:sft_data}
\end{table}

\scalebox{0.9}{$\bullet$} \textit{ASR}: Existing datasets are used as shown in Table~\ref{tab:training_data}. An example prompt is "Transcribe this audio".

\scalebox{0.9}{$\bullet$} \textit{Audio Caption}: This task involves generating audio descriptions using the AudioCaps test set \citep{audiocaps}, with English references translated into Thai for Thai Audio Captioning. The evaluation metric is METEOR.

\scalebox{0.9}{$\bullet$} \textit{Speech Translation}: Thai-to-English is from CommonVoice (Thai), and target English texts are derived from translation. English/X-to-Thai is from Covost2, and target Thai texts are derived from translating English texts. X refers to a non-English audio language (Arabic, German, Spanish, French, Indonesian, Italian, Japanese, Chinese) taken from Covost2. The translation was performed using our internal system, which matches Google Translate API performance. An example prompt is "Translate this audio into \texttt{language}". Each setup includes 2000 examples. The evaluation metric is BLEU.

\scalebox{0.9}{$\bullet$} \textit{Gender Classification}: Fleurs is used as gender labels are available for both English and Thai. The metric is accuracy. 

\scalebox{0.9}{$\bullet$} QA examples in SALMONN/LTU are based on short spoken documents (under 10 seconds). To enable longer audio understanding, we segmented Yodas2~\citep{li2023yodas} audio into 30-second chunks and used GPT-4o to generate QA pairs, including multiple-choice questions (MCQs) to improve SpokenQA performance (Section~\ref{section:audio_sft}). We focused on the Thai subset of Yodas2 to address the dominance of English in existing QA datasets. Additionally, we generated QA pairs from the VISTEC-SER dataset~\citep{vistec_ser}, leveraging metadata like speaker gender and emotional state to capture voice-specific characteristics.

% We create Yodas2-30S-QA 
% Given speech as context, this task involves answering textual questions related to the speech context. For English, SpokenSQuAD \citep{lee2018spoken} is used as an English benchmark. 

% For Thai, we take speech utterance with duration longer than 20 seconds from CommonVoice (resulting in 665 QA pairs), we use GPT-4 to generate questions and answers based on text references. The metric is F1 token overlap.

\scalebox{0.9}{$\bullet$} \textit{Audio Caption}: AudioCaps is used for pre-training, but its short ground-truth captions limit detailed response generation. To address this, we provide Gemini-1.5-Pro with both audio input and the short caption, prompting it to generate detailed responses. This augmented data is called AudioCaps (Gemini).

\begin{figure}[!h]
    \centerline{
\includegraphics[width=0.7\linewidth,keepaspectratio]{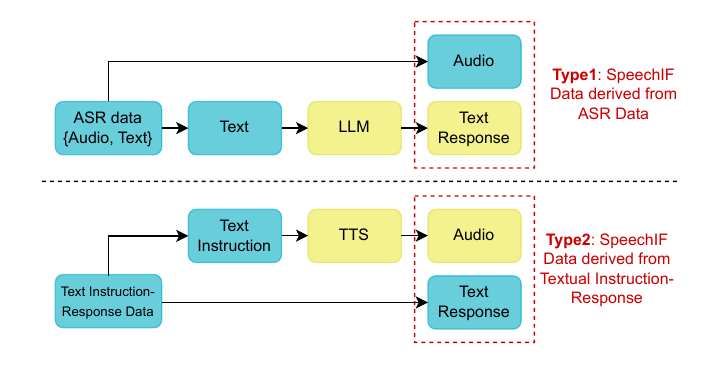}}
    \caption{Speech Instruction Following Data Creation Pipeline}
    \label{fig:speech_if}
\end{figure}

\scalebox{0.9}{$\bullet$} \textit{Speech Instruction Following (SpeechIF)}: This task requires models to listen to spoken instructions and directly respond. Current models like SALMONN lack specific data for this ability. We propose two methods for generating SpeechIF data (Figure~\ref{fig:speech_if}). \textit{Type1} leverages ASR datasets to generate text responses from transcripts. However, since ASR data typically contains non-question utterances, LLMs often default to safe responses such as ``I'm sorry, as an AI assistant I cannot..." in up to 30\% of cases. While it offers voice diversity, it does not fully reflect real-world interactions. \textit{Type2} synthesizes speech from instruction-response pairs (e.g., Alpaca, Airoboros), providing more practical commands but struggles with unsuitable instructions like math or coding. Though lacking voice diversity, it represents real interactions better. For evaluation, we selected instructions from AlpacaEval (English) and SelfInstruct (Thai), creating SpeechIF benchmark for both languages. The prompt for baseline models (e.g., SALMONN) is ``Listen to the audio and answer the question". 

\scalebox{0.9}{$\bullet$} \textit{Complex Instruction Following (ComplexIF)}: We propose ComplexIF to assess models' ability to follow unseen, compound instructions, where each instruction involves two to three audio tasks (such as transcribe, then translate). In ComplexIF, models have to respond in specific formats (e.g., JSON, XML), with format templates provided in the instruction prompt. As it evaluates the general instruction following ability, only English speech data is used. ComplexIF is used exclusively for evaluation, without additional training.

\subsubsection{Experimental Setup}

\noindent \textbf{Evaluation}: For existing tasks, we use standard metrics. For SpeechIF and ComplexIF, we follow MT-Bench \citep{mtbench} in using an LLM judge (GPT-4o), and we adapt the single-turn evaluation prompt from MT-Bench and score responses on a scale from 1.0 to 10.0. For ComplexIF, we prompt the judge to evaluate the response on two aspects: 

\textit{(1) Quality} considers helpfulness, relevance, accuracy, depth, creativity, and level of detail of the response.

\textit{(2) Format} considers how well the response follows the format required by the user (e.g., JSON, XML, Markdown, etc).

\noindent \textbf{Baselines}: Competitive audio language models include,

\textit{(1) Open-weights}: Qwen-Audio (Qwen-7B) \citep{chu2023qwen}, SALMONN (Vicuna-13B) \citep{tang2024salmonn}, and DiVA (Llama3-8B) \citep{held2024diva}. For these open models, we use available weights on HuggingFace.

\textit{(2) Proprietary}: Gemini-1.5-Pro (Audio) \texttt{gemini-1.5-pro-001} through Google API with \{audio, text\_instruction\} as input.

\subsubsection{Results and Findings}
\label{section:speech_encoding_results}
\textit{Remarks}: The majority of experiments in Section~\ref{section:speech_encoding_results} were conducted prior to the development of the Typhoon2 Text series. Thus, the best performing 8B Typhoon backbone at the time, Typhoon1.5 "\texttt{llama-3-typhoon-v1.5-8b-instruct}", was selected as the main LLM backbone. In the final experiment (discussed in Finding 4), we provide the performance of the new audio model with Typhoon2 "\texttt{typhoon-2-llama-31-8b-instruct-beta-v1}".

\subsubsection*{Finding 1: Performance Disparities of Audio Language Models in English vs. Thai}
The results in Table~\ref{tab:results_part1} and Table~\ref{tab:results_part2} demonstrate that: (1) Baselines using multilingual backbones exhibit significant performance degradation in Thai, while Gemini-1.5-Pro maintains strong performance across both Thai and English. (2) Among the baselines, DiVA is the only model that performs well on the SpeechIF task, but it experiences a notable drop when tested on Thai. Thus, the subsequent experiments aim to develop a model that can effectively handle these tasks in both English and a low-resource language such as Thai.

% \subsubsection*{Finding: Should the audio encoder be trained?}
% \begin{table}[!h]
%     \centering
%     \begin{tabular}{lcc}
%     \toprule
%         \textbf{Trainable Weights}    &\textbf{Test-clean}  &\textbf{Test-other}  \\ 
%         \midrule
%         Adapter Only         &6.31  &9.65  \\
%         Adapter + AudioEncoder     &4.49  &11.43 \\
%     \bottomrule
%     \end{tabular}
%     \caption{The results when training only on LibriSpeech }
%     \label{tab:audio_backbone}
% \end{table}
% While all existing works trained the LoRA weights of the language model, SALMONN \citep{tang2024salmonn} and LTU-AS \citep{gong2023ltuas} kept the audio encoder unchanged and trained the adapter. In contrast, Qwen-Audio \citep{chu2023qwen} trained the audio encoder without having an adapter. Based on the architecture in Figure~\ref{fig:audio_typhoon2audio}, we first investigate whether the audio encoder should also be trained. By training on LibriSpeech, the results in Table~\ref{tab:audio_backbone} show that training the audio encoder alongside the adapter does not yield a clear improvement. Given the added complexity and extended training time, subsequent experiments will, thus, proceed with the audio encoder \textit{frozen}.

\subsubsection*{Finding 2: Pre-training Speech Encoder and LLM Backbones}
\label{section:exp_pretraining}
This experiment focuses on selecting backbones, and comparing Whisper with its English+Thai fine-tuned variant. Similarly, Typhoon is a Llama-3 model fine-tuned to English+Thai. Our results (in Table~\ref{tab:pretraining_results}) show that for ASR, models where both backbones are matched with the target language yield the best results. However, for audio captioning, the performance difference between these models is marginal. As a low-resource language such as Thai is our goal, Whisper+Th coupled with Typhoon-1.5 are selected.
% for subsequent experiments.

\begin{table}[!h]
    \centering
    \begin{tabular}{llcccc}
    \toprule
        \multicolumn{2}{c}{\textbf{Backbone}}  &\multicolumn{2}{c}{\textbf{ASR (WER$\downarrow$)}} &\multicolumn{2}{c}{\textbf{AC (METEOR$\uparrow$)}} \\
        \textbf{Speech} &\textbf{LLM} &\textbf{En} &\textbf{Th*} &\textbf{En} &\textbf{Th}  \\ 
        \midrule
        Whisper-v3-large        &Llama-3      &\textbf{6.02} &16.66 &30.75 &20.04  \\
        Whisper-v3-large        &Typhoon-1.5           &7.76 &20.01 &29.56 &\textbf{20.62}  \\
        Whisper-v3-large-Th     &Llama-3      &7.35 &15.68 &29.52 &19.94  \\
        Whisper-v3-large-Th     &Typhoon-1.5           &9.15 &\textbf{13.52} &\textbf{30.83} &20.55  \\
    \bottomrule
    \end{tabular}
    \caption{Pre-training Results on ASR: LibriSpeech (other), CommonVoice (*subset-1K), AC: AudioCaps (En\&Th-translated) }
    \label{tab:pretraining_results}
\end{table}

\subsubsection*{Finding 3: Recipe for Supervised Fine Tuning (SFT) Data Mixture}
\label{section:audio_sft}
This experiment focuses on data mixture to enhance instruction-following abilities across tasks and languages. Training is initialized using the pre-trained model from the previous section. The results in Table~\ref{tab:sft_results} show that: \textit{First}, the pre-trained model does not exhibit task ability and it simply provides transcriptions of speech regardless of instructions. \textit{Second}, when fine-tuned on only English prompt-response pairs (a subset of around 600K pairs in total taken from SALMONN and LTU), the model achieves better performance on new tasks, but performs poorly on Thai ASR, showing similar characteristics to SALMONN in Table~\ref{tab:results_part1}. Ultimately, our SFT recipe significantly improves the model performance on evaluated tasks, while not significantly degrading its ASR abilities. Further information on our SFT recipe is provided our the Typhoon-Audio paper \citep{manakul2024enhancing}.

\begin{table}[!h]
\tabcolsep=1.4mm
    \centering
    \begin{tabular}{lcccccc}
    \toprule
       \textbf{Experiment}  &\textbf{\#Ex} &\textbf{ASR*$\downarrow$} &\textbf{Th2En$\uparrow$} &\textbf{SpokenQA$\uparrow$} &\textbf{SpeechIF$\uparrow$} &\textbf{ComplexIF$^\dagger$$\uparrow$} \\        
        \midrule 
        Pre-trained  &- &\textbf{13.52} &0.00 &28.33 &1.12 &1.41 \\ % 1.741, 1.075  
        100\% English SFT  &600K  &80.86 &6.01  &36.88 &1.48 &6.35 \\ % 5.025, 7.683 % v2: SALMONN + LTU | 600k English QA only
        Our SFT recipe     &640K  &16.89 &\textbf{24.14} &\textbf{64.60} &\textbf{6.11} &\textbf{7.54} \\
    \bottomrule
    \end{tabular}
    \caption{SFT Results on Thai Tasks and English ComplexIF.  $^*$ASR is eval on subset-1K of CV17. $^\dagger$Average of Qual and Format. For 100\% English SFT, around 600K QA pairs were taken from SALMONN and LTU datasets.}
    \label{tab:sft_results}
\end{table}

\newpage
\subsubsection*{Finding 4: Typhoon-Audio \& Typhoon2-Audio versus Existing Audio Language Models}
% -------------------------------------------------------------------------------- %

\textit{Typhoon-Audio}: 
We evaluate our Typhoon-Audio model, based on the Typhoon 1.5 LLM, against competitive benchmarks (Tables~\ref{tab:results_part1} and~\ref{tab:results_part2}). For ASR, Typhoon-Audio is one of two models (along with Gemini) achieving a WER below 15.0 on Thai ASR, despite underperforming in English. In translation, it surpasses SALMONN and Gemini 1.5 Pro in Thai-to-English, demonstrating strong Thai comprehension and English generation. For voice characteristics, Typhoon-Audio performs comparably to SALMONN in English-to-Thai gender recognition. In spoken document QA, it matches Gemini 1.5 Pro, making it the only open-source model for Thai QA. For speech instruction following, Typhoon-Audio outperforms Gemini-1.5-Pro in both English and Thai and approaches Gemini 1.5 Pro in handling complex instructions. It also has a lower hallucination rate than prior models~\citep{sun2024crosscheckgpt}, though hallucination in speech instruction remains a challenge.

\textit{Typhoon2-Audio}: As noted at the beginning, most experiments were conducted before the development of the Typhoon 2 LLM. Once available, we applied the same pre-training and SFT methodologies, using the same speech backbones, to create Typhoon2-Audio. Results in Tables~\ref{tab:results_part1} and~\ref{tab:results_part2} show improved performance over Typhoon-Audio in ASR, speech translation, spoken QA, and speech instruction following. However, Typhoon2-Audio underperforms in gender classification and occasionally fails to respond to spoken instructions during complex tasks, despite strong question-answering capabilities.

\begin{table}[!h]
    \centering
    \begin{tabular}{lcccccccc}
    \toprule
        \multirow{2}{*}{\textbf{Model}} &\multirow{2}{*}{\textbf{Size}} &\multicolumn{2}{c}{\textbf{ASR (WER$\downarrow$)}} &\multicolumn{3}{c}{\textbf{Translation (BLEU$\uparrow$)}}  &\multicolumn{2}{c}{\textbf{Gender (Acc$\uparrow$)}}  \\ 
        & &\textbf{En} &\textbf{Th} &\textbf{Th2En} &\textbf{En2Th} &\textbf{X2Th} &\textbf{En} &\textbf{Th} \\
        \midrule
        Qwen-Audio     &7B     &6.94  &95.12 &0.00 &2.48 &0.29 &37.09 &67.97 \\
        SALMONN        &13B    &\textbf{5.79}  &98.07 &14.97 &0.07  &0.10  &95.69 &93.26 \\
        DiVA           &8B     &30.28 &65.21 &7.97  &9.82  &5.31  &47.30 &50.12 \\
        Gemini-1.5-Pro &-  &5.98  &\textbf{13.56} &22.54 &20.69 &13.52 &90.73 &81.32 \\
        \midrule
        Typhoon-Audio  &8B     &8.72  &14.17 &24.14 &17.52 &10.67 &\textbf{98.76} &\textbf{93.74} \\
        Typhoon2-Audio         &8B     &5.83  &14.04  &\textbf{33.25}  &\textbf{27.15}   &\textbf{15.93}   &76.51 &75.65   \\
    \bottomrule
    \end{tabular}
    \caption{Audio LM Evaluation in English and Thai on ASR, Translation, Gender Classification. Size refers to the size of the LLM.}
    \label{tab:results_part1}
\end{table}

\begin{table}[!h]
    \tabcolsep=1.5mm
    \centering
    \begin{tabular}{lcccccccc}
    \toprule
        \multirow{2}{*}{\textbf{Model}} &\multirow{2}{*}{\textbf{Size}} &\multicolumn{2}{c}{\textbf{SpokenQA (F1$\uparrow$)}} &\multicolumn{2}{c}{\textbf{SpeechIF (Judge$\uparrow$)}}  &\multicolumn{3}{c}{\textbf{ComplexIF (Judge$\uparrow$)}}  \\ 
        & &\textbf{En} &\textbf{Th} &\textbf{En} &\textbf{Th} &\textbf{Qual} &\textbf{Format} &\textbf{Avg.} \\
        \midrule
        Qwen-Audio    &7B     &25.34 &0.00 &1.07 &1.03 &3.13 &1.68 &2.41 \\
        SALMONN       &13B    &52.92 &2.95  &2.47 &1.18  &4.10 &5.09 &4.60\\
        DiVA          &8B     &44.52 &15.13 &\textbf{6.81} &2.68 &6.33 &7.83  &7.08 \\
        Gemini-1.5-Pro &-  &\textbf{74.09} &62.10 &3.24 &3.93 &\textbf{7.25} &{8.99} &\textbf{8.12} \\
        \midrule
        Typhoon-Audio &8B     &48.83 &64.60 &5.62 &6.11 &6.34 &8.73 &7.54 \\
        Typhoon2-Audio         &8B    &69.22 &\textbf{70.01} &6.00 &\textbf{6.79}  &5.35$^\dagger$ &\textbf{9.01}  &7.18 \\
    \bottomrule
    \end{tabular}
    \caption{Audio LM Evaluation in English and Thai on Spoken QA, Speech Instruction Following, and English Complex Instruction Following. Size refers to the size of the LLM. $^\dagger$We observed examples where Typhoon2-Audio did not provide any answer to speech instruction in nested commands; hence, receiving very low scores on these examples.}
    \label{tab:results_part2}
\end{table}

% -------------------------------------------------------------------------------- %

%% file: 5.2_audio_output.tex
\newpage
\subsection{Speech Generation}
\label{section:speech_generation}
The speech encoding components (Section~\ref{section:speech_encoder}) and LLM enable processing of audio, speech, and text inputs to produce text outputs. For speech output, the generated text can be passed to a text-to-speech (TTS) system, but this pipelined approach delays TTS until text generation finishes, causing high time-to-first-token latency. This section explores extending the output to enable parallel speech and text generation.

\begin{figure}[!h]
    \centerline{
\includegraphics[width=\linewidth,keepaspectratio]{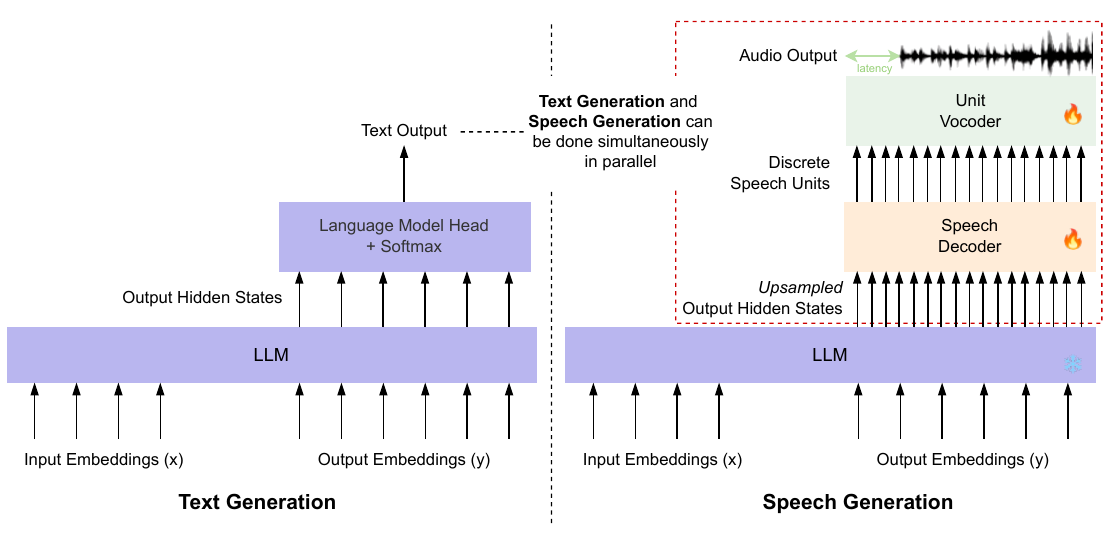}}
    \caption{Text Generation (left) and Speech Generation (right)}
    \label{fig:speech_generation}
\end{figure}

We extend the output side for speech generation using the Llama-Omni architecture \citep{fang2024llama}, as shown in Figure~\ref{fig:speech_generation}. Hidden states from the LLM are upsampled and fed into a non-autoregressive speech decoder, producing discrete speech units. A unit vocoder then maps these tokens to a waveform.

% The training process is divided into two stages. In the first stage, the speech decoder is trained using Connectionist Temporal Classification (CTC) \citep{graves2006connectionist} to predict target discrete speech units. Separately, the second stage focuses on training the unit vocoder, which is trained using combinations of losses such as (i) reconstruction loss, which measures the difference between generated and target waveforms (e.g., L1, L2, or spectral loss); (ii) adversarial loss, which encourages naturalness by training a discriminator to distinguish real from synthesized waveforms; (iii) feature matching loss, which minimizes differences between real and synthesized waveforms in the discriminator's intermediate feature space; (iv) Duration prediction loss, which minimizes the mean square error (MSE) between the predicted and ground truth durations of each unit segment in the logarithmic domain, enhancing the temporal alignment of the generated speech.

The training process has two stages. First, the speech decoder is trained with Connectionist Temporal Classification (CTC) \citep{graves2006connectionist} to predict discrete speech units. Second, the unit vocoder is trained using multiple losses: (1) reconstruction loss (e.g., L1, L2, or spectral) for waveform accuracy, (2) adversarial loss for naturalness, (3) feature matching loss to align intermediate features, and (4) duration prediction loss (MSE) to improve temporal alignment of unit segments.

\subsubsection{Speech Decoder}
The LLM encodes output tokens, $y_{1:t}$, into their hidden representations:
\begin{equation}
    \mathbf{h}_{1:t} = f(y_{1:t} ; \boldsymbol{\theta}_{\texttt{LLM}})
\end{equation}
To perform text generation, these hidden representations $\mathbf{h}_{1:t}$ are passed to a linear layer, mapping them into the vocabulary space. To perform speech generation, $\mathbf{h}_{1:t}$ are passed to a speech decoder, which can operate in parallel with text generation.

% Scale-up
Since a speech waveform requires a higher number of discrete tokens compared to their corresponding textual form, the hidden representations is upsampled by a factor of $\lambda$ to obtain the input of the speech decoder:

\begin{equation}
    \mathbf{h}'_{1:\lambda \times t} = [\underbrace{\mathbf{h}_{1},...,\mathbf{h}_{1}}_{\lambda}, \underbrace{\mathbf{h}_{2},...,\mathbf{h}_{2}}_{\lambda}, ..., \underbrace{\mathbf{h}_{t},...,\mathbf{h}_{t}}_{\lambda} ]    
\end{equation}
where upsampling factor, $\lambda$, is set to 25. Next, upsampled hidden representations $\mathbf{h}'$ are passed to the speech decoder, which is implemented with a stack of causal decoder layers (e.g., 4 Llama's layers in this work) and a feedforward network. This speech decoder is causal but non-autoregressive in both training and inference. Subsequently, the outputs of the speech decoder $\mathbf{o}_{1:\lambda \times t}$ are aligned with the target discrete speech units using CTC.

\subsubsection{Discrete Speech Tokens}
The discrete unit based architecture follows Llama-Omni \citep{fang2024llama} and SpeechGPT \citep{zhang2023speechgpt}. However, instead of using HuBERT \citep{hsu2021hubert}, this work adopts XEUS \citep{chen-etal-2024-towards-robust} to generate discrete speech tokens. This is because XEUS is a model designed to support multiple languages, which is crucial for efficient speech synthesis and vocoder applications. Essentially, XEUS converts continuous a speech waveform into discrete speech units. These discrete units correspond to the number of clusters ($K$) from the k-means clustering algorithm applied to XEUS representations. The centroids or k-means vectors are trained on 20\% of {28.7K} audio examples randomly selected from \textit{Mix-1} by using the TTS system (described in Section 5.3.4).

\subsubsection{Unit Vocoder}
A unit vocoder \citep{polyak21_interspeech}, based on the HiFi-GAN architecture \citep{kong2020hifi}, takes a sequence of self-supervised discrete representations (in our case, they are k-means of XEUS tokens) to resynthesize speech. By disentangling essential components of speech—such as linguistic content, prosodic features, and speaker identity—into distinct low-bitrate representations, it enables efficient and high-quality speech synthesis. In this work, the unit vocoder is employed to generate speech outputs from discrete units.

\subsubsection{Data}
\label{section:speech_decoder_data}

As developing a single-speaker speech generation system is simpler than a multi-speaker one, this work focuses on single-speaker output. However, due to limited single-speaker Thai TTS data, we use the Google Cloud Platform TTS system (\texttt{th-TH-Standard-A}) to synthesize speech waveforms from textual data. The synthesized data are derived from:

\scalebox{0.9}{$\bullet$} \textit{Mix-1}: This data mixture includes 28.7K examples derived from Thai self-instruct (8.7K) and a translated Alpaca subset (20K). Responses are regenerated using GPT-4o-mini in a conversational style, following Llama-Omni's setup.\footnote{It was intended to fine-tune the speech encoder + LLM for more conversational responses, but experiments showed it degraded text generation, so the original speech encoder + LLM was retained.}

\scalebox{0.9}{$\bullet$} \textit{Mix-2}: This mixture comprises 220K examples derived from: (1) 75K re-written Thai SFT data,\footnote{Dataset sourced from \url{https://huggingface.co/datasets/Suraponn/thai_instruction_sft}.} (2) responses generated from re-written instructions, and (3) 70K examples of sentences using Thai unique names (e.g., companies, locations).\footnote{Thai company names were limited, so this portion was upsampled.} Instructions and responses follow the same GPT-4o-mini conversational generation style as Mix-1 but on a larger scale.

\scalebox{0.9}{$\bullet$} \textit{Mix-3}: This mixture focuses on diversity and contains 155K examples. It includes: (1) responses from Mix-2 SFT data (excluding instructions), (2) Thai unique name data (without upsampling), (3) 21K English Alpaca examples rewritten conversationally and LJSpeech for English TTS, (4) code-mixed sentences with Thai sentences containing English words, and (5) generated sentences involving numbers (e.g., phone numbers, dates, years). , with Table~\ref{tab:speech_output_data} provides the summary.

\begin{table}[!ht]
    \centering
    \begin{tabular}{lc}
    \toprule
    \textbf{Dataset}  &\textbf{\#Examples}   \\
    \midrule
    Thai Response     & 75,120  \\
    Thai Unique Names & 12,787  \\
    Alpaca (English)  & 21,816  \\
    LJSpeech          & 13,100  \\
    Thai-English Mix  & 21,000  \\
    Number            & 11,550  \\
    \midrule
    \textbf{Total}             & 155,371 \\
    \bottomrule
    \end{tabular}
    \caption{The final data mixture (Mix-3) for speech decoder.}
    \label{tab:speech_output_data}
\end{table}

\subsubsection{Experimental Setup}

\textbf{Evaluation}: In speech generation (similar to TTS), we evaluate the quality of generated speech on two aspects: {Accuracy} and {Naturalness} as follows:

\scalebox{0.9}{$\bullet$} \textit{Accuracy}: The generated speech was transcribed using Whisper-v3-large-turbo as the ASR system, and the character error rate (CER) was computed by comparing the transcribed text to the original text.

\scalebox{0.9}{$\bullet$} \textit{Naturalness}: The UTokyo-SaruLab MOS (UTMOS) system \citep{saeki2022utmos}, developed for the VoiceMOS Challenge 2022 \citep{huang2022voicemos}, is a state-of-the-art tool for predicting speech quality using Mean Opinion Scores (MOS) from 1 (poor) to 5 (excellent). It should be noted that while UTMOS performs well across diverse contexts, its accuracy declines with non-English speech, highlighting the need for improvements to better handle language-specific features and support multilingual environments.

\textbf{Baselines}: 

\scalebox{0.9}{$\bullet$} When evaluating Typhoon2-Audio-as-TTS, we benchmark it against systems, including, (1) \textit{Open-source}: (1.1) PyThaiTTS \citep{pythaitts}, Thai text-to-speech model based on Coqui-TTS trained on TSync-1 and TSync-2 data; (1.2) Seamless (\texttt{seamless-m4t-v2-large}) \citep{seamless2023}, a unified multilingual system that can synthesize Thai speech among many languages; (1.3) MMS-TTS (\texttt{facebook/mms-tts}) \citep{pratap2023scalingspeechtechnology1000}, massively multilingual speech project, aiming to provide speech technology across a diverse range of languages. (2) \textit{Proprierary}: (2.1) Google Cloud Platform (GCP) \_TTS (\texttt{th-TH-Standard-A}) and (2.2) Microsoft Azure TTS (Premwadee) through their APIs. 

\scalebox{0.9}{$\bullet$}  When evaluating Typhoon2-Audio for end-to-end speech-to-speech tasks, we benchmark it against existing end-to-end models, including Llama-Omni (open-source) and GPT-4o-Audio (proprietary through API).

\subsubsection{Results and Findings}

\subsubsection*{Finding 1: Developing a Thai unit vocoder}

We evaluated several unit vocoder models trained on different data mixtures. One model consistently outperformed the others, providing superior synthesis quality, even when trained on a large dataset. However, models trained on larger datasets did not always perform better. In some cases, performance declined, suggesting that too much diverse data might complicate the learning process. Although we could not identify a concrete reason for this, further investigation may be needed.

Model selection was primarily based on qualitative evaluation. We assessed audio quality through perceptual listening and coherence checks, selecting the model that produced the most natural, high-fidelity audio. While subjective, this method effectively identified the best model.
This evaluation highlights the need to balance data diversity with performance and to use flexible criteria for model selection.

\subsubsection*{Finding 2: Data Mixture for Speech Decoder}

Here, we investigate different data mixes (described in Section~\ref{section:speech_decoder_data}) for speech decoder. The vocoder (trained on data Mix-2) is fixed, and we train only the speech decoder for around 8 epochs. The results in Table~\ref{tab:tts_cer} and Table~\ref{tab:tts_utmos} show that data Mix-3 yields the best overall accuracy and naturalness.

\begin{table}[!ht]
    \centering
    \begin{tabular}{lccccc}
    \toprule
    \textbf{Training Data} & \textbf{Overall (1k)} & \textbf{En+Th} & \textbf{Name} & \textbf{General-Th} & \textbf{Number}  \\
    \midrule
    % \multicolumn{6}{l}{Baseline} \\
    % PyThaiTTS          & 81.74 & 92.39 & 70.00 & 81.53 & 63.04 \\
    % Seamless           & 27.90 & 33.40 & 22.12 & 23.90 & 41.68 \\
    % GCP\_TTS           & 12.64 & 23.76 & 12.44 & 7.13 & 6.61  \\
    % Azure\_Premwadee   & 12.28 & 23.50 & 11.92 & 6.58 & 7.26  \\
    % \rowcolor{Gray}
    % \multicolumn{6}{l}{Typhoon2-Audio (Speech Decoder's Variant)} \\
    Mix-1 & 21.29 & 38.23 & 19.26 & 11.73 & 22.04 \\ % path4
    Mix-2 & 20.15 & 34.15 & 19.43 & 11.88 & 21.43 \\ % path3
    Mix-3 (base)        & 18.76 & 28.71 & 19.72 & 11.93 & 23.28 \\ % path1
    + LJSpeech          & 19.35 & 28.65 & 18.62 & \textbf{11.05} & 37.87 \\ % path0
    + LJSpeech + Number & \textbf{18.27} & \textbf{28.19} & \textbf{17.73} & 13.16 & \textbf{14.65} \\ % path2
    \bottomrule
    \end{tabular}
    \caption{Character Error Rate (CER) $\downarrow$ of synthesized speech across different categories.}
    \label{tab:tts_cer}
\end{table}

\begin{table}[!ht]
    \centering
    \begin{tabular}{lccccc}
    \toprule
    \textbf{Training Data} & \textbf{Overall (1k)} & \textbf{En+Th} & \textbf{Name} & \textbf{General-Th} & \textbf{Number}  \\
    \midrule
    % PyThaiTTS          & 2.7938 & 2.9544 & 2.6312 & 2.7148 & 2.8699 \\
    % Seamless           & 3.7143 & 3.8199 & 3.6056 & 3.6744 & 3.7056 \\
    % GCP\_TTS           & 3.6036 & 3.6435 & 3.6171 & 3.5742 & 3.6169 \\
    % Azure\_Premwadee   & 4.0557 & 4.0697 & 3.9848 & 4.0532 & 4.0975 \\
    % \rowcolor{Gray}
    % \multicolumn{6}{l}{Typhoon2-Audio (Speech Decoder's Variant)} \\
    Mix-1               & 3.05 & 2.93 & 3.06 & \textbf{3.13} & 3.06 \\ % path4
    Mix-2               & 3.08 & 3.02 & 3.04 & \textbf{3.13} & 3.07 \\ % path3
    Mix-3 (base)        & 3.10 & 3.11 & 3.04 & \textbf{3.13} & 3.01 \\ % path1
    + LJSpeech          & 3.11 & 3.14 & 3.07 & 3.12 & 2.98 \\ % path0
    + LJSpeech + Number & \textbf{3.13} & \textbf{3.16} & \textbf{3.12} & \textbf{3.13} & \textbf{3.08} \\ % path2
    \bottomrule
    \end{tabular}
    \caption{Objective Quality Assessment (UTMOS) $\uparrow$ scores across different categories}
    \label{tab:tts_utmos}
\end{table}

\subsubsection*{Finding 3: Typhoon2-Audio as Text-to-Speech}
As Typhoon2-Audio can take text as input without audio or text inputs, the model (LLM + speech decoder + unit vocoder) is capable of synthesizing speech from raw text. This means that Typhoon2-Audio can act as a \textbf{text-to-speech (TTS)} system. It should be noted that using Typhoon2-Audio for TTS is entirely non-autoregressive. This experiment investigates the TTS performance and compares it with existing TTS systems. 

As shown in Table~\ref{tab:tts_cer2} and Table~\ref{tab:tts_utmos2}, Typhoon2-Audio-as-TTS achieves the lowest CER, below 20\%, on the overall (1k) subset among open TTS models. In terms of naturalness, while Seamless attains a higher UTMOS score, its synthesized English speech resembles native English speakers speaking Thai. Conversely, Typhoon2-Audio-as-TTS produces English speech that sounds more like native Thai speakers speaking English.

\begin{table}[!ht]
    \centering
    \tabcolsep=0.75mm
    \begin{tabular}{lcccccc}
    \toprule
    \textbf{System} & \textbf{Type} & \textbf{Overall (1k)} & \textbf{En+Th} & \textbf{Name} & \textbf{General-Th} & \textbf{Number}  \\
    \midrule
    PyThaiTTS        & Open   & 81.74 & 92.39 & 70.00 & 81.53 & 63.04 \\
    Seamless         & Open     & 27.90 & 33.40 & 22.12 & 23.90 & 41.68 \\
    MMS-TTS          & Open     & 27.50 & 38.04 & 25.32 & 18.44 & 48.51 \\
    GCP\_TTS         & Proprietary     & 12.64 & 23.76 & 12.44 & 7.13 & \textbf{6.61}  \\
    Azure\_Premwadee & Proprietary     & \textbf{12.28} & \textbf{23.50} & \textbf{11.92} & \textbf{6.58} & 7.26  \\
    \midrule
    Typhoon2-Audio-as-TTS & Open & 18.27 & 28.19 & 17.73 & 13.16 & 14.65 \\ % path2
    \bottomrule
    \end{tabular}
    \caption{Character Error Rate (CER) $\downarrow$ of synthesized speech across different categories.}
    \label{tab:tts_cer2}
\end{table}

\begin{table}[!ht]
    \centering
    \tabcolsep=0.6mm
    \begin{tabular}{lcccccc}
    \toprule
    \textbf{System} & \textbf{Type} & \textbf{Overall (1k)} & \textbf{En+Th} & \textbf{Name} & \textbf{General-Th} & \textbf{Number}  \\
    \midrule
    PyThaiTTS        & Open  & 2.79 & 2.95 & 2.63 & 2.72 & 2.87 \\
    Seamless         & Open  & 3.71 & 3.82 & 3.61 & 3.67 & 3.71 \\
    MMS-TTS         & Open  & 3.71 & 3.74 & 3.55 & 3.73 & 3.66 \\
    GCP\_TTS         & Proprietary  & 3.60 & 3.64 & 3.62 & 3.57 & 3.62 \\
    Azure\_Premwadee & Proprietary  & \textbf{4.06} & \textbf{4.07} & \textbf{3.99} & \textbf{4.05} & \textbf{4.10} \\
    \midrule
    Typhoon2-Audio-as-TTS & Open & 3.13 & 3.16 & 3.12 & 3.13 & 3.08 \\ % path2
    \bottomrule
    \end{tabular}
    \caption{Objective Quality Assessment (UTMOS) $\uparrow$ scores across different categories}
    \label{tab:tts_utmos2}
\end{table}

\newpage
\subsection{End-to-End Speech-to-Speech Evaluation}

Typhoon2-Audio's ability to generate text and speech responses from spoken instructions is evaluated through speech-to-text (S2TIF) and speech-to-speech (S2SIF) tasks, with a focus on S2SIF here. Two aspects are assessed: \textbf{content generation} and \textbf{speech quality}. Content generation is evaluated using the LLM-as-a-judge framework, while speech quality is measured by accuracy (e.g., CER, WER) and naturalness (e.g., UTMOS).

Spoken instructions are taken from SpeechIF (English and Thai splits) \citep{manakul2024enhancing}, using the prompt: "\textit{Respond conversationally to the speech provided in the language it is spoken in}", similar to Talk Arena \citep{talkarena2024}. For content evaluation, automatic speech recognition (ASR) (Gemini-1.5-Flash) transcribes the generated speech. The LLM judge (GPT-4o) assesses the transcript on two aspects:
\begin{itemize}
    \item \textbf{Quality}: helpfulness, relevance, accuracy, depth, creativity.
    \item \textbf{Style}: suitability in a conversational setting.
\end{itemize}

The results in Table~\ref{tab:endtoeend_speech1} show that Typhoon2-Audio outperforms Llama-Omni for English but falls short of GPT-4o-Audio. For Thai, Typhoon2-Audio significantly outperforms Llama-Omni, which responds in English to Thai inputs, and performs competitively with GPT-4o-Audio. All models show reduced performance on transcribed speech due to imperfections in speech generation, with Typhoon2-Audio showing a larger drop in English, as its speech generation is optimized for Thai. Despite this, transcribed scores suggest the generated speech remains usable.

\begin{table}[!ht]
    \centering
    \begin{tabular}{lcccc}
    \toprule
        \multirow{2}{*}{\textbf{Model}}  &\multicolumn{2}{c}{\textbf{SpeechIF (English)}}  &\multicolumn{2}{c}{\textbf{SpeechIF (Thai)}} \\
        &\textbf{Quality($\uparrow$)} &\textbf{Style($\uparrow$)}   &\textbf{Quality($\uparrow$)} &\textbf{Style($\uparrow$)}   \\
        \midrule
        % SpeechGPT             \\
        %% Results using Text output
       \rowcolor{Gray}
        \multicolumn{5}{l}{\textbf{Results using Text Output}} \\
        Llama-Omni      &5.58 &6.52   &1.88  &2.53     \\ % &9.11B param, &10.0, &1.23
        GPT-4o-Audio    &\textbf{7.23} &\textbf{8.25}   &6.96  &\textbf{8.38}     \\ % &7.90, &9.93
        Typhoon2-Audio  &6.34 &7.12   &\textbf{7.43}  &8.18     \\ % &9.98, &9.88
       \rowcolor{Gray}
        \multicolumn{5}{l}{\textbf{Results using Transcribed Speech}} \\
        Llama-Omni     & 5.15 & 5.79 & 1.71 & 2.14  \\
        GPT-4o-Audio   & \textbf{6.82} & \textbf{7.86} & 6.66 & \textbf{8.07}  \\
        Typhoon2-Audio & 4.92 & 5.39 & \textbf{7.19} & 8.04  \\
        % \midrule
    \bottomrule
    \end{tabular}
    \caption{S2TIF Evaluation of end-to-end systems.}
    \label{tab:endtoeend_speech1}
\end{table}

We evaluate end-to-end systems on speech generation quality, measuring transcription error rates and UTMOS scores, similar to Section~\ref{section:speech_generation}.

As shown in Table~\ref{tab:endtoeend_speech2}, Typhoon2-Audio performs comparably to GPT-4o-Audio in Thai but requires improvement in English. While Llama-Omni achieves the lowest WER and CER for Thai, it responds exclusively in English, unlike the other models.

Typhoon2-Audio also shows lower UTMOS scores than GPT-4o and Llama-Omni, indicating its speech output is perceived as less natural.

% TODO: revise the numbers in Table below and write a story

\begin{table}[!ht]
    \centering
    \begin{tabular}{lcccccc}
    \toprule
        \multirow{2}{*}{\textbf{Model}}  &\multicolumn{3}{c}{\textbf{SpeechIF (English)}}  &\multicolumn{3}{c}{\textbf{SpeechIF (Thai)}} \\
        &\textbf{WER($\downarrow$)} &\textbf{CER($\downarrow$)} &\textbf{UTMOS($\uparrow$)}  &\textbf{WER($\downarrow$)} &\textbf{CER($\downarrow$)} &\textbf{UTMOS($\uparrow$)}  \\
        \midrule
        % SpeechGPT             \\
        Llama-Omni      & 4.98  &  3.40 & \textbf{3.932} &  {8.51}$^\dagger$ &  {6.30}$^\dagger$    & {3.928}$^\dagger$  \\
        GPT-4o-Audio    & \textbf{4.88} &   \textbf{3.20}  & 3.652 &   11.71   &  8.05    &  3.464  \\
        Typhoon2-Audio  & 33.00 &   26.50  & 2.285 &   10.04   &  8.67  &  2.348  \\
        % \midrule
    \bottomrule
    \end{tabular}
    \caption{Speech Quality: End-to-End Evaluation (without ASR or TTS systems). $^\dagger$Llama-Omni fails to respond in Thai. As Llama-Omni simply responds in English, its responses achieve good results in Thai SpeechIF as measured by WER, CER, UTMOS.}
    \label{tab:endtoeend_speech2}
\end{table}